\pdfoutput=1
\documentclass[10pt,journal,compsoc]{IEEEtran}
\usepackage{amsmath,amsfonts}
\usepackage{algorithmic}
\usepackage{algorithm}
\usepackage{array}
\usepackage[caption=false,font=normalsize,labelfont=sf,textfont=sf]{subfig}
\usepackage{textcomp}
\usepackage{stfloats}
\usepackage{url}
\usepackage{verbatim}
\usepackage{graphicx}
\usepackage[backref]{hyperref}

\ifCLASSOPTIONcompsoc
  \usepackage[nocompress]{cite}
\else
  \usepackage{cite}
\fi

\begin{document}

\title{From Linguistic Giants to Sensory Maestros: A Survey on Cross-Modal Reasoning with Large Language Models}

\author{
        Shengsheng~Qian,~\IEEEmembership{Member,~IEEE}, %
        Zuyi~Zhou,
        Dizhan~Xue,
        Bing~Wang,
        and~Changsheng~Xu,~\IEEEmembership{Fellow,~IEEE}
\IEEEcompsocitemizethanks{

  \IEEEcompsocthanksitem Shengsheng~Qian, Zuyi~Zhou, and Dizhan~Xue are with State Key Laboratory of Multimodal Artificial Intelligence Systems, Institute of Automation, Chinese Academy of Sciences, Beijing 100190, China, and also with University of Chinese Academy of Sciences (e-mail: shengsheng.qian@nlpr.ia.ac.cn; zhouzuyi2023@ia.ac.cn; xuedizhan17@mails.ucas.ac.cn). \\
  Bing~Wang is with School of Computer Science and Engineering, Tianjin University of Technology, Tianjin 300384, China. (e-mail: wb502@stud.tjut.edu.cn).\\
  Changsheng~Xu is with State Key Laboratory of Multimodal Artificial Intelligence Systems, Institute of Automation, Chinese Academy of Sciences, Beijing 100190, China, University of Chinese Academy of Sciences, and Peng Cheng Laboratory. (e-mail: csxu@nlpr.ia.ac.cn).
}
\thanks{
(Corresponding author: Changsheng Xu.)}
}

\markboth{}%
{}


\IEEEtitleabstractindextext{
\begin{abstract}
Cross-modal reasoning (CMR), the intricate process of synthesizing and drawing inferences across divergent sensory modalities, is increasingly recognized as a crucial capability in the progression toward more sophisticated and anthropomorphic artificial intelligence systems. Large Language Models (LLMs) represent a class of AI algorithms specifically engineered to parse, produce, and engage with human language on an extensive scale. The recent trend of deploying LLMs to tackle CMR tasks has marked a new mainstream of approaches for enhancing their effectiveness. This survey offers a nuanced exposition of current methodologies applied in CMR using LLMs, classifying these into a detailed three-tiered taxonomy. Moreover, the survey delves into the principal design strategies and operational techniques of prototypical models within this domain. Additionally, it articulates the prevailing challenges associated with the integration of LLMs in CMR and identifies prospective research directions. To sum up, this survey endeavors to expedite progress within this burgeoning field by endowing scholars with a holistic and detailed vista, showcasing the vanguard of current research whilst pinpointing potential avenues for advancement. An associated GitHub repository that collects the relevant papers can be found at \href{https://github.com/ZuyiZhou/Awesome-Cross-modal-Reasoning-with-LLMs}{https://github.com/ZuyiZhou/Awesome-Cross-modal-Reasoning-with-LLMs}.
\end{abstract}

}

\IEEEdisplaynontitleabstractindextext

\maketitle

\section{Introduction}

\IEEEPARstart{I}{n} recent years, the advent of Large Language Models (LLMs) has revolutionized the field of artificial intelligence, showcasing unprecedented capabilities in natural language understanding and generation. These models, such as GPT-4 \cite{openai2023gpt4}, Gemini \cite{team2023gemini}, Claude \cite{anthropic2024claude}, and Llama 2 \cite{touvron2023llama}, have demonstrated remarkable proficiency in tasks ranging from text completion to question answering. 
Despite the great success of LLMs, the community has also acknowledged some critical limitations in their capabilities.
Some work indicates that LLMs have constraints in understanding the physical world and performing complex reasoning \cite{valmeekam2022large,wang2023can,laban2023summedits}. 
Furthermore, the concept of extending powerful LLMs to multimodal models has gained traction within the realm of multimodal learning \cite{ye2023mplug,zheng2023minigpt,liu2024visual}.
As a result, recent developments in methodologies for Cross-Modal Reasoning (CMR) with LLMs have shown significant progress alongside notable challenges.

\begin{figure}[ht]
    \centering
    \includegraphics[width=0.8\linewidth]{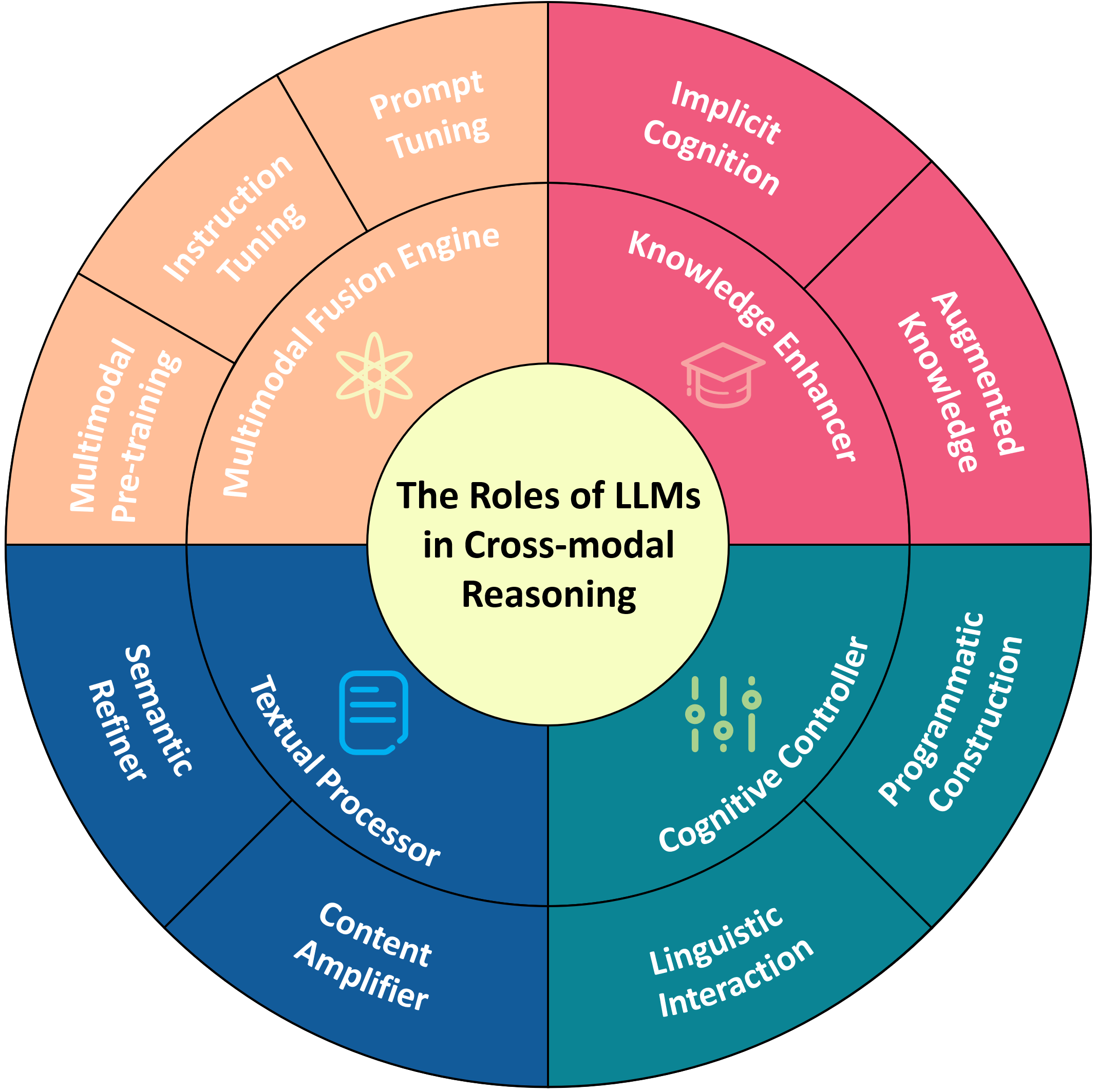}
    \caption{The taxonomy of the roles of LLMs in cross-modal reasoning.}
    \label{fig:tox1}
\end{figure}

\begin{figure*}[t]
    \centering
    \includegraphics[width=0.9\linewidth]{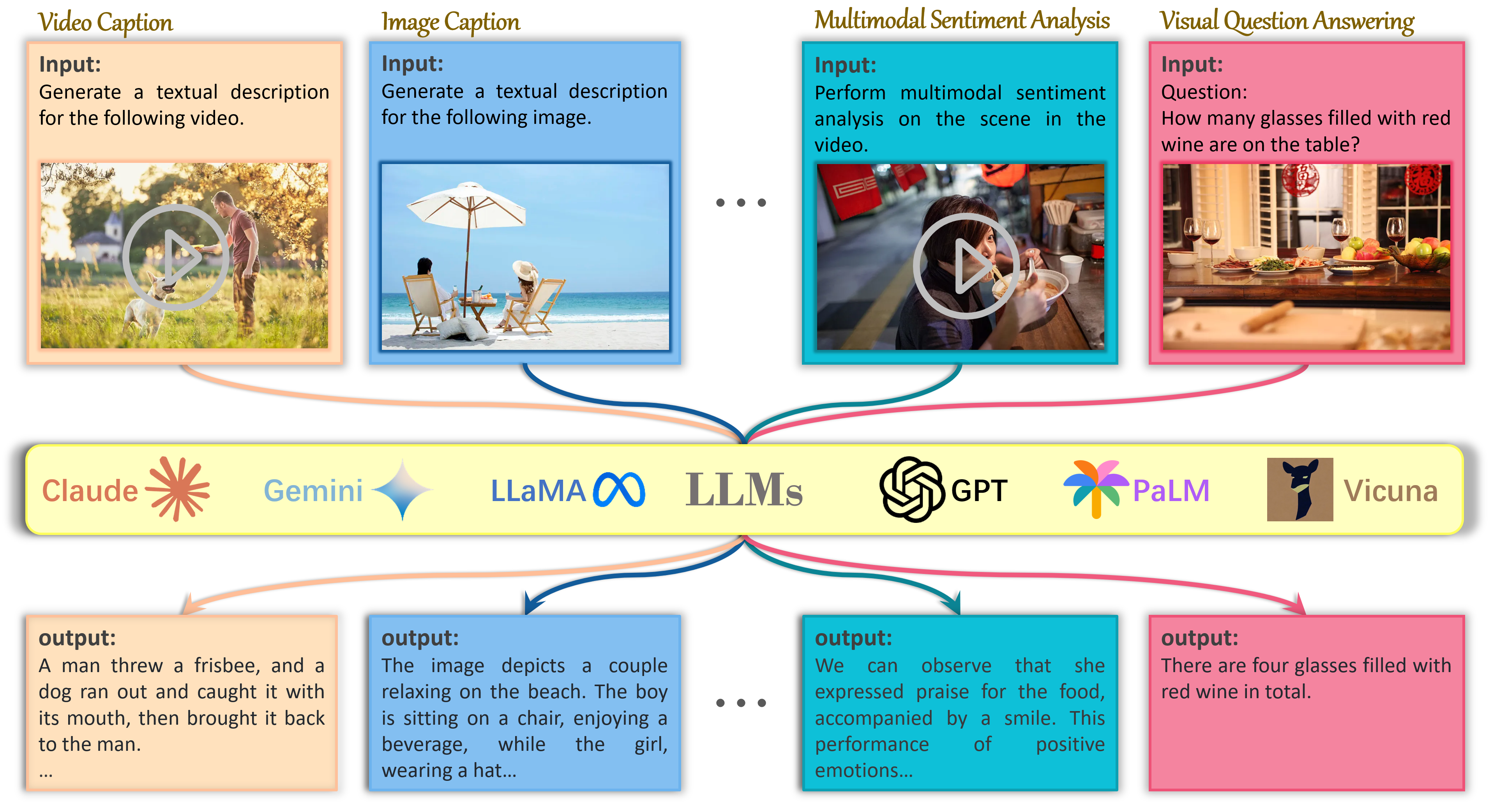}
    \caption{Examples of Cross-Modal Reasoning utilizing Large Language Models (LLMs). The illustrated scenarios highlight the potent capabilities of LLMs in facilitating effective cross-modal reasoning and aiding in comprehension across various modalities.}
    \label{fig:o}
\end{figure*}

Cross-Modal Reasoning with Large Language Models (CMR with LLMs) integrates the capacity of LLMs to comprehend and infer new information across diverse modalities, such as text, image, and sound. 
Reasoning is a cognitive process that entails a systematic analysis and assessment of information utilizing logical structures, critical thinking principles, and empirical evidence to formulate informed conclusions, facilitate decision-making processes, and resolve complex problems \cite{sloman1996empirical}.
In particular, CMR \cite{xue2023survey,malkinski2022review} encompasses the comprehensive analysis and reasoning across various modalities. Through harnessing the interplay and interdependencies among multiple modalities, CMR aims to extract significant insights and make coherent inferences.
In the context of LLMs, this capability is derived from their extensive training on diverse datasets, allowing them to make connections and generate responses that span different types of data.
The development of CMR has profoundly impacted and transformed a multitude of specialized domains. This progress embraces an array of multimodal tasks, including visual question answering \cite{Agrawal2015VQAVQ, Hudson2019GQAAN}, vision-and-language navigation \cite{anderson2018vision, hao2020towards}, image captioning \cite{Vinyals2014ShowAT, Xu2015ShowAA}, video captioning \cite{Luo2020UniViLMAU, Huang2020MultimodalPF}, etc. 
Fig. \ref{fig:o} illustrates several examples showcasing the application of LLMs in CMR.

As shown in Fig. \ref{fig:tox1}, methods for CMR with LLMs can be roughly classified into four groups according to the roles of LLMs in CMR, as follows:
(1)\textbf{ LLM as multimodal fusion engi}\textbf{}\textbf{ne}: LLMs act as integrative frameworks that consolidate and harmonize data from diverse modalities. Through sophisticated algorithms and neural network architectures, these models \cite{zhu2023minigpt,zheng2023minigpt,zhao2023chatbridge} discern and interpret the multifaceted semantics of tasks, enabling a comprehensive understanding that transcends the limitations of single-modality processing.
(2) \textbf{LLM as textual processor}: LLMs apply advanced natural language processing techniques to distill and extract critical information from extensive textual data, which enables the extraction of responses from voluminous textual data \cite{zeng2022socratic, zhu2023pointclip}. Moreover, LLMs can enrich and optimize textual expression for CMR, such as enriching answers \cite{You2023IdealGPTID} or prompts \cite{guo2023viewrefer, zhu2022pointclip}.
(3) \textbf{LLM as cognitive controller}: LLMs architect the reasoning trajectory, orchestrating the sequence of cognitive steps required to address complex tasks. This involves generating structured plans, reasoning pathways \cite{yang2023mm, huang2022inner}, or even executable code \cite{singh2023progprompt, gupta2023visual} that guide the model towards the resolution of tasks.
(4) \textbf{LLM as knowledge enhancer}: LLMs enrich tasks with implicit knowledge from extensive pre-training while also integrating explicit domain-specific information through external databases or auxiliary tools. LLMs may produce textual representations of visual content \cite{guo2023viewrefer} or facilitate auxiliary modules \cite{zhang2023prompt, pratt2023does} for enhanced content comprehension and reasoning. Additionally, LLMs can extract pertinent information from web sources or search engines \cite{yang2023mm, shen2023hugginggpt}.

To the best of our knowledge, this is the first survey that provides an investigation into the roles of LLMs in facilitating CMR.
To clarify a comprehensive context for our analysis, we furnish a brief overview of published surveys partially related to our work: 
Some surveys \cite{berger2017spatio,zhao2023survey} vigorously examine the nature of LLMs, thoroughly categorizing and elucidating their distinct attributes. 
Some other surveys \cite{huang2022towards, wang2023aligning, qiao2022reasoning}, meanwhile, elaborate on the reasoning methodologies associated with LLMs, even though their inquiries do not encompass the multimodal perspective.
Concentrating primarily on multimodal component incorporation strategies, \cite{yin2023survey} reviews the state-of-the-art landscape of multimodal LLMs. 
Different from the above surveys, we review extensive research on Cross-Modal Reasoning with LLMs to provide a comprehensive and up-to-date survey.
%
The objective of this survey is to comprehensively outline the contemporary state-of-the-art developments in CMR with LLMs. Our contribution can be succinctly summarized as follows:
\begin{enumerate}

\item{\textbf{Innovative Taxonomy:} We introduce a novel taxonomy for existing approaches for CMR with LLMs, providing a systematic understanding of mixed elements and methodologies in this interdisciplinary domain.}

\item{\textbf{Comprehensive Analysis:} We undertake a thorough exploration of the existing models and research in CMR with LLMs, incorporating diverse perspectives, methodologies, and tasks, thus delivering a comprehensive synopsis of current advancements in the field.}

\item{\textbf{Extensive Resources:} We collect and discuss abundant resources on CMR with LLMs, including state-of-the-art methods and practical tasks. We hope this survey can serve as a valuable resource for anyone interested in the intersection of cross-modal reasoning and large language models.}

\item{\textbf{Discussion of Challenges and Future Directions:} We highlight the existing challenges in CMR with LLMs and then sketch potential future paths and research opportunities. We aim to kindle novel research ideas and foster growth in this burgeoning interdisciplinary field.}

\end{enumerate}

\begin{figure*}
    \centering
    \includegraphics[width=0.9\linewidth]{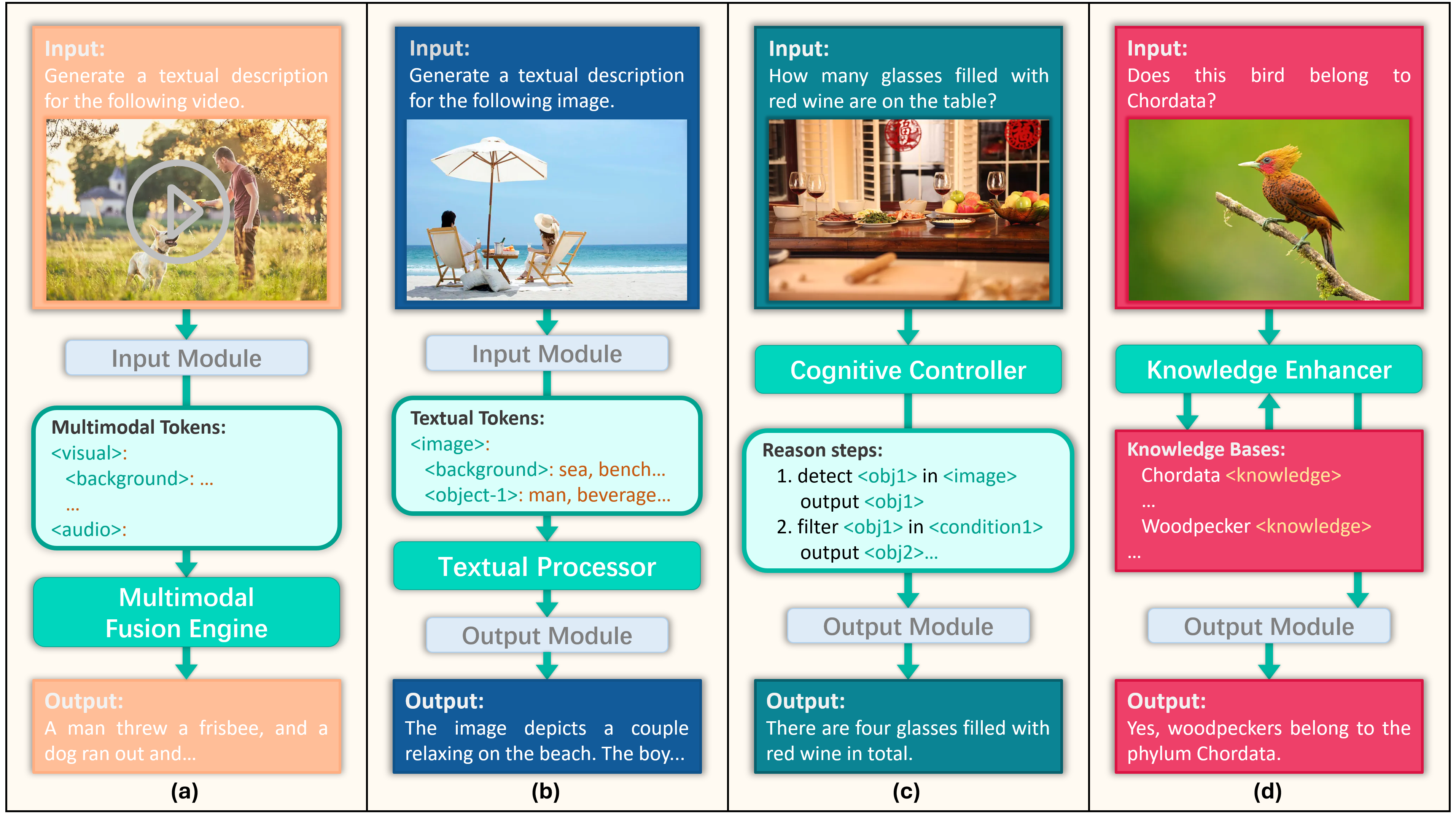}
    \caption{The multifaceted role of Large Language Models (LLMs) within the domain of Cross-Modal Reasoning (CMR): 
    (a) LLMs as multimodal fusion engines enact a pivotal function in the alignment, fusion, and integration of multimodal inputs into coherent textual representations. These multimodal inputs necessitate a series of transformations to render them compatible with the intricate architecture of LLMs. Subsequently, the processed data is relayed to auxiliary output modules, culminating in the generation of responses. (b) LLMs as textual processors analyze different types of textual tokens and generate appropriate texts to fulfill the requirements of other modules. (c) LLMs as cognitive controllers exercise a critical evaluative role in discerning the task-specific requirements and appraising the practical feasibility of potential implementations, thereby orchestrating the reasoning methodology. The reasoning sequences produced by LLMs are subsequently operationalized by supplementary modules in the system. (d) LLMs as knowledge enhancers can provide valuable support to CMR tasks by offering a wealth of knowledge derived not just from their large datasets but also from external sources and real-time contextual information.}
    \label{fig:roles}
\end{figure*}

\section{Taxonomy for Cross-Modal Reasoning with Large Language Models}

As illustrated in Fig. \ref{fig:tox1}, we present a taxonomy of the implementation approaches for cross-modal reasoning (CMR) utilizing large language models (LLMs). From the perspective of the roles of LLMs in CMR, these methods are roughly classified into four categories: Multimodal Fusion Engine, Textual Processor, Cognitive Controller, and Knowledge Enhancer.
\begin{figure*}
    \centering
    \includegraphics[width=1\linewidth]{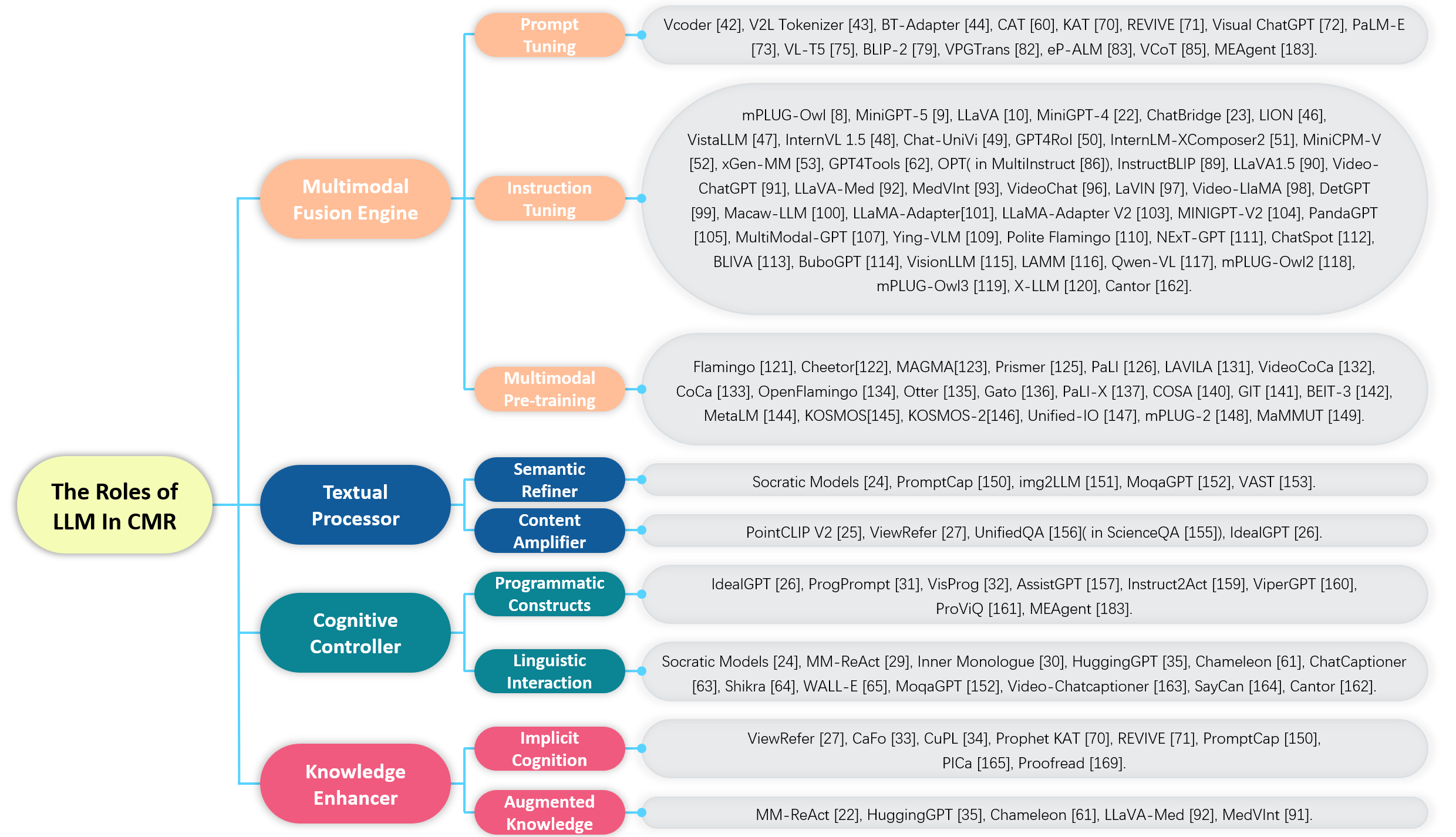}
    \caption{Comprehensive overview of Cross-Modal Reasoning with Large Language Models (CMR with LLMs) categories and corresponding methodologies. }
    \label{fig:modellsit}
\end{figure*}
\subsection{Multimodal Fusion Engine}
LLMs exhibit the capacity to effectively integrate natural language with diverse modalities, including visual or audio inputs. This integration enhances their reasoning abilities, enabling them to furnish more nuanced and informed responses. A Multimodal Fusion Engine (MFE) guarantees the precision of these responses by meticulous mapping of multimodal features into a language-aligned feature space, thus mitigating the risk of generating inaccurate answers or hallucinations. By adopting this integrated approach, LLMs harness insights from various heterogeneous sources, enriching their comprehension and reasoning capabilities, thereby culminating in responses that are more sophisticated and comprehensive.

MFE plays a crucial role in integrating multimodal data to facilitate cross-modal reasoning. It serves as a fundamental component of LLMs, bridging the gap between various modalities. Presently, three commonly employed methods have been identified to accomplish this integration: Prompt Tuning, Instruction Tuning, and Pre-training.

\begin{itemize}
    \item \textbf{Prompt Tuning:} The specialization in prompt tuning involves the careful design and calibration of tailored textual prompts to guide LLMs toward specific operational areas. Precision in crafting these prompts is vital as it aims to evoke desired responses or behaviors from the models. In the realm of prompt tuning for LLMs, there are two main categories based on prompt formatting: discrete prompt tuning and continuous prompt tuning. Discrete prompt tuning involves creating and using distinct text templates or natural language prompts, while soft prompt tuning utilizes continuous vectors or embeddings as the foundation for prompts. Currently, a combination of these two types of prompts is often employed in CMR tasks. At the heart of prompt tuning lies the transformation of visual information into tokens to construct effective prompts \cite{jain2024vcoder,zhu2024beyond, liu2024bt}.
    
    \item \textbf{Instruction Tuning:} In the context of LLMs, instruction tuning \cite{liu2024visual, zhang2023multi} refers to the practice of adjusting and improving the pedagogical approach utilized for these models. The goal is to enhance the models' understanding of specific directives provided by users. This typically involves clarifying and making task descriptions more precise, or rewording directives in a manner that helps the model accurately interpret the desired outcome. Models can improve performance on cross-modal reasoning (CMR) tasks by fine-tuning instructions from multiple perspectives. This includes adjusting granularity levels \cite{chen2024lion, pramanick2024jack, chen2024far}, optimizing region-level visual-language combinations \cite{jin2024chat, zhang2023gpt4roi, dong2024internlm}, or integrating both approaches \cite{yao2024minicpm, xue2024xgen}.
    
    \item \textbf{Multimodal Pre-training:} Significant progress has been made in the development of large-scale multimodal pre-training foundation models, driven by innovative pre-training objectives and model architectures. These foundation models primarily leverage several approaches for multimodal tasks, including next token prediction \cite{radford2018improving, NEURIPS20201457c0d6, radford2019language}, masked language modeling \cite{devlin2018bert, liu2019roberta}, and encoder-decoder structures \cite{devlin2018bert, liu2019roberta, lewis2019bart}.
\end{itemize}

\subsection{Textual Processor}

The profound linguistic and semantic expertise possessed by LLMs is harnessed to enhance textual processing, particularly within cross-modal reasoning tasks. In such contexts, modules often generate textual data to facilitate communication between components. Due to the varying designs of these modules, the generated textual content may not always align seamlessly with the specific tasks' requirements. The role of the Textual Processor (TP) is to refine and harmonize this textual information, transforming it into coherent and contextually appropriate natural language expressions or generating customized text suited to precise needs \cite{zeng2022socratic}. By prioritizing consistency and fluency, LLMs demonstrate exceptional proficiency in integrating diverse information sources and producing text that meets a wide range of contextual and functional requirements \cite{Wang2023CaptionAI, zhang2023prompt, zhu2022pointclip}.

\begin{itemize}
    \item \textbf{Semantic Refiner:} As semantic refiners, LLMs leverage their language capabilities to refine and distill textual information. In these cases, an expansive corpus of textual data from diverse, heterogeneous sources is provided as input to the LLMs. LLMs then meticulously process and seamlessly integrate this textual compendium, leveraging its sophisticated language understanding to generate concise, semantically refined output that succinctly encapsulates the key conceptual essence distilled from the broader informational corpus.
    \item \textbf{Content Amplifier:}The capabilities of LLMs can be harnessed to expand and enrich textual information. By leveraging their advanced language generation abilities, LLMs can produce supplementary text that offers enhanced contextual framing, detailed descriptions, and illustrative examples. This enriched output serves as valuable input for subsequent models or as fully developed answers tailored to specific tasks.
\end{itemize}

\subsection{Cognitive Controller}

Large Language Models (LLMs) act as Cognitive Controllers (CC), coordinating reasoning processes by generating inputs for various inference tools. They serve as a bridge between the language model and other tools, ensuring seamless integration for proficient reasoning and inference generation. This enhances cognitive processing and resource utilization.

To understand LLMs as Cognitive Controllers, we explain their role in cross-modal reasoning. LLMs break down complex tasks into manageable parts and delegate them to suitable computational tools or modules \cite{shen2023hugginggpt, lu2023chameleon, yang2023gpt4tools, gupta2023visual}. They can create strategic outlines or identify specific modules for engagement, primarily using programming constructs and natural language interfaces.

\begin{itemize}
    \item \textbf{Programmatic Construction:} LLMs convert cross-modal reasoning tasks into structured programs, thereby enabling the precision crafting of reasoning sequences and facilitating the generation of standardized responses. Notably, the application VisProg \cite{gupta2023visual} utilizes GPT-3 to create a visual script, in which each line of code is meticulously mapped to a module that addresses a designated sub-task. Additionally, LLMs can also be tasked with the provision of parameter identifiers for inputs requisite for module functionality, employing crafted contextual exemplars as benchmarks to adeptly maneuver through complexities \cite{shen2023hugginggpt}.
    
    \item \textbf{Linguistic Interaction:} Contrasting with the rigidity of programmatic directives, natural language presents a more dynamic modality for interaction, enabling models to curate responses that are both broader in scope and logically coherent. Typically, these models formulate reasoning pathways either by generating elaborative monologues \cite{huang2022inner, zeng2022socratic} or by orchestrating contextual dialogues with peer models, which facilitate a seamless flow of reasoning \cite{Zhu2023ChatGPTAB, chen2023shikra, wang2023wall}. This modality allows for a nuanced and flexible approach to cross-modal reasoning that is adaptable to emergent complexities and user expectations.
\end{itemize}

\subsection{Knowledge Enhancer}
LLMs can amalgamate and condense information derived from extensive knowledge bases, thus enabling them to generate outputs rich in informative content. This capability is attained through a methodical, iterative process \cite{You2023IdealGPTID}, wherein LLMs conduct searches within the knowledge base, appraise the contextual cues and historical data, assess the adequacy of the available information to fulfill tasks or address queries, and ultimately present the response in a user-friendly manner.

Cross-modal reasoning pertains to the integration of data from various modalities, such as text, images, and audio, to derive coherent and meaningful deductions. Nonetheless, accomplishing successful cross-modal reasoning often necessitates additional knowledge beyond what is presently available.
In this context, LLMs assume a pivotal role as Knowledge Enhancers (KE). LLMs represent formidable AI models that have undergone extensive training on copious amounts of textual data, exhibiting human-like language generation capabilities. Employing these models enables the supplementation of existing knowledge bases with augmented contextual information, thereby facilitating cross-modal reasoning.
According to the source of provided knowledge, these methods can be divided into two groups: Implicit Cognition and Augmented Knowledge.

\begin{itemize}
    \item \textbf{Implicit Cognition:} Within the realm of CMR tasks, LLMs are employed directly, harnessing the extensive knowledge acquired during their comprehensive training phases. This deployment capitalizes on the LLMs’ sophisticated language comprehension abilities, derived from their exposure to a vast corpus of textual data, to facilitate reasoning across diverse modalities including text, imagery, and auditory inputs. Notably, LLMs can generate textual interpretations of visual content \cite{guo2023viewrefer} or provide prompts \cite{zhang2023prompt, pratt2023does} that support auxiliary modules in their processing tasks, thereby enhancing comprehension and reasoning about the multimodal content.

    \item \textbf{Augmented Knowledge:} LLMs also serve as reservoirs of external knowledge, augmenting CMR tasks with additional information that enriches the existing knowledge base. In this capacity, LLMs function as dynamic tools for the retrieval or generation of pertinent textual content, effectively supplementing the information pool available for CMR undertakings. This might involve leveraging LLMs to source relevant textual data from digital repositories, websites, or Search Engines \cite{yang2023mm, shen2023hugginggpt}, thereby broadening the scope of knowledge accessible for informed reasoning and analysis.
\end{itemize}

\section{LLM as Multimodal Fusion Engine}

\begin{figure}
    \centering
    \includegraphics[width=0.8\linewidth]{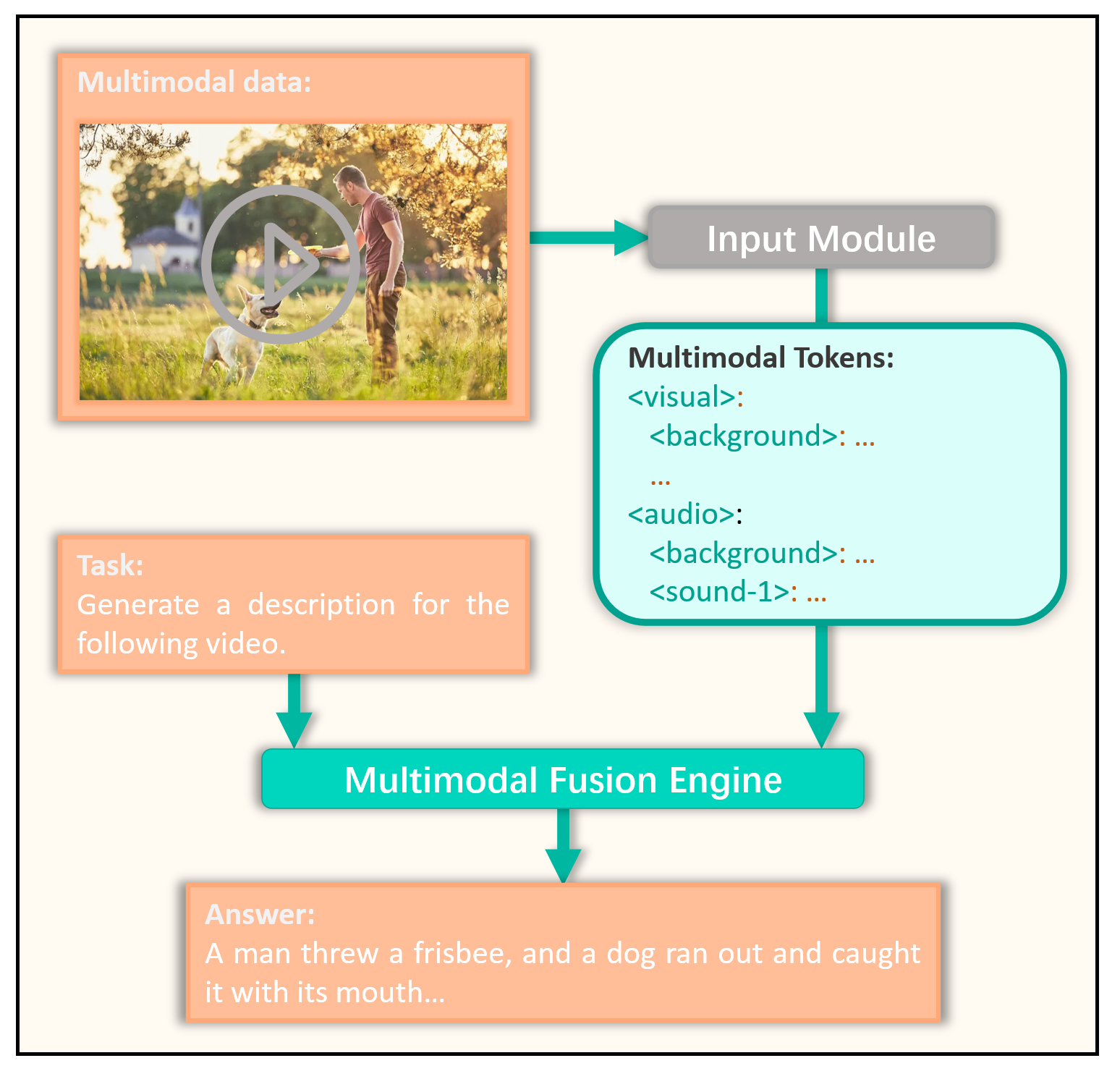}
    \caption{The function of Large Language Models (LLMs) as Multimodal Fusion Engine within cross-modal reasoning frameworks.}
    \label{fig:exp-mfe}
\end{figure}

Fig. \ref{fig:exp-mfe} illustrates the utilization of LLMs as a Multimodal Fusion Engine(MFE) within the context of cross-modal reasoning. A typical CMR task encompasses a textual task description accompanied by multimodal data inputs. The transformation of multimodal data into an array of tokens, which serve as the LLMs' inputs, is critical for achieving LLM-based information fusion. These tokens are frequently comprised of specialized prompts, along with code or machine languages that are easily interpretable by LLMs. The LLMs are adept at extracting vast amounts of textual information and refining it to formulate a concise answer. 

\begin{figure}
    \centering
    \includegraphics[width=0.9\linewidth]{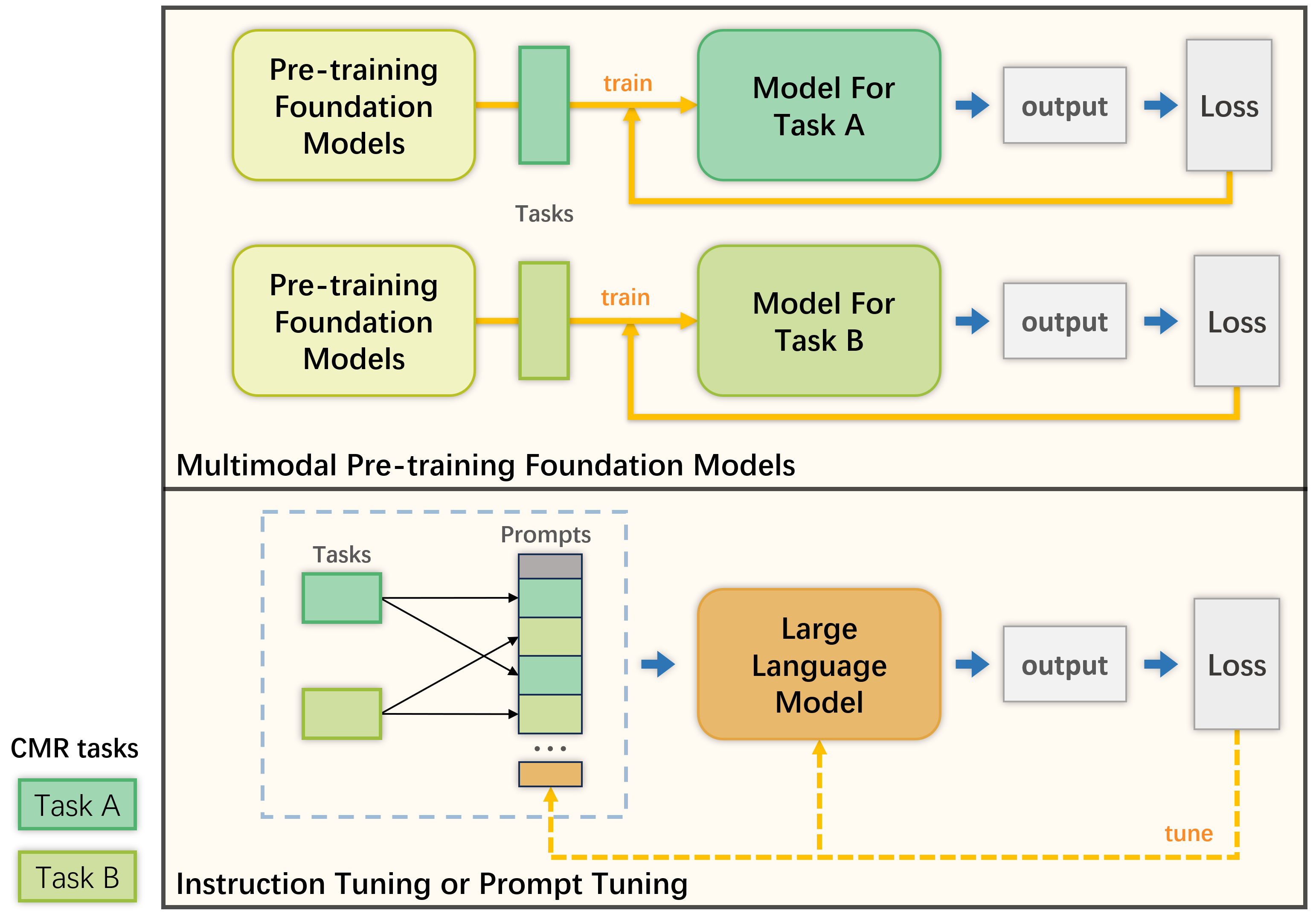}
    \caption{Two categories of approaches for cross-modal reasoning with large language models are identified: Prompting and Multimodal Pre-training. Prompting refines prompts for models while Multimodal Pre-training uses CMR data to train foundational models.}
    \label{fig:tox2}
\end{figure}

\begin{figure*}
    \centering
    \includegraphics[width=0.9\linewidth]{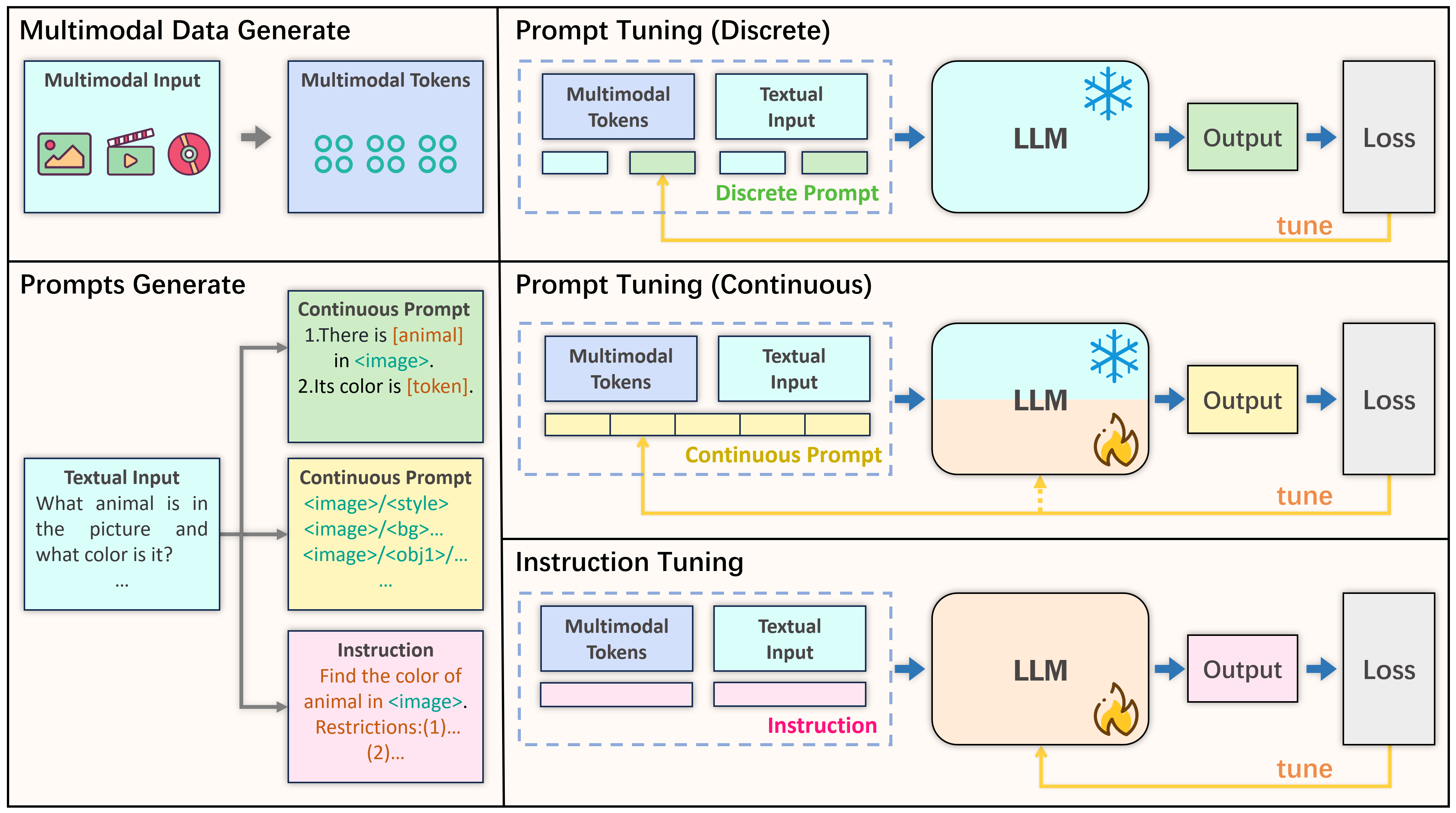}
    \caption{The caption delineates sequential illustrations of the Large Language Models (LLMs) prompting procedure. Initially, the models synthesize multimodal data derived from various multimodal inputs. This is then followed by the conversion of input tasks into a spectrum of prompts, as indicated on the left of the illustration. These prompts, in concert with the multimodal data and the initial input tasks, are subsequently fed into a Large Language Model. The output from this model is textual data, which is then used to compute a loss function that facilitates the fine-tuning of the prompts, a process illustrated on the right. It is noteworthy to mention that only instruction-tuned LLMs have the ability to generalize to unseen tasks through the utilization of new instructions.}
    \label{fig:prompt-t}
\end{figure*}

The current MFE can be categorized into three methods: prompt tuning, instruction tuning, and multimodal pre-training. On a broader scale, models that utilize prompt tuning and instruction tuning for training their modules follow a specific prompting process. For a clearer understanding, we utilize Fig.\ref{fig:tox2} to illustrate the distinction between prompting and pre-training. Additionally, we leverage the differences in tuning to delineate between the two major categories of prompting methods. The development of effective prompts is a focal point in various research endeavors, often employing prompt tuning—a method involving the creation and refinement of prompt tokens \cite{10.1145/3560815}. This procedure can be categorized into prompt tuning and instruction tuning (outlined in Fig. \ref{fig:prompt-t}).

\subsection{Prompt Tuning}

\begin{figure}
    \centering
    \includegraphics[width=1\linewidth]{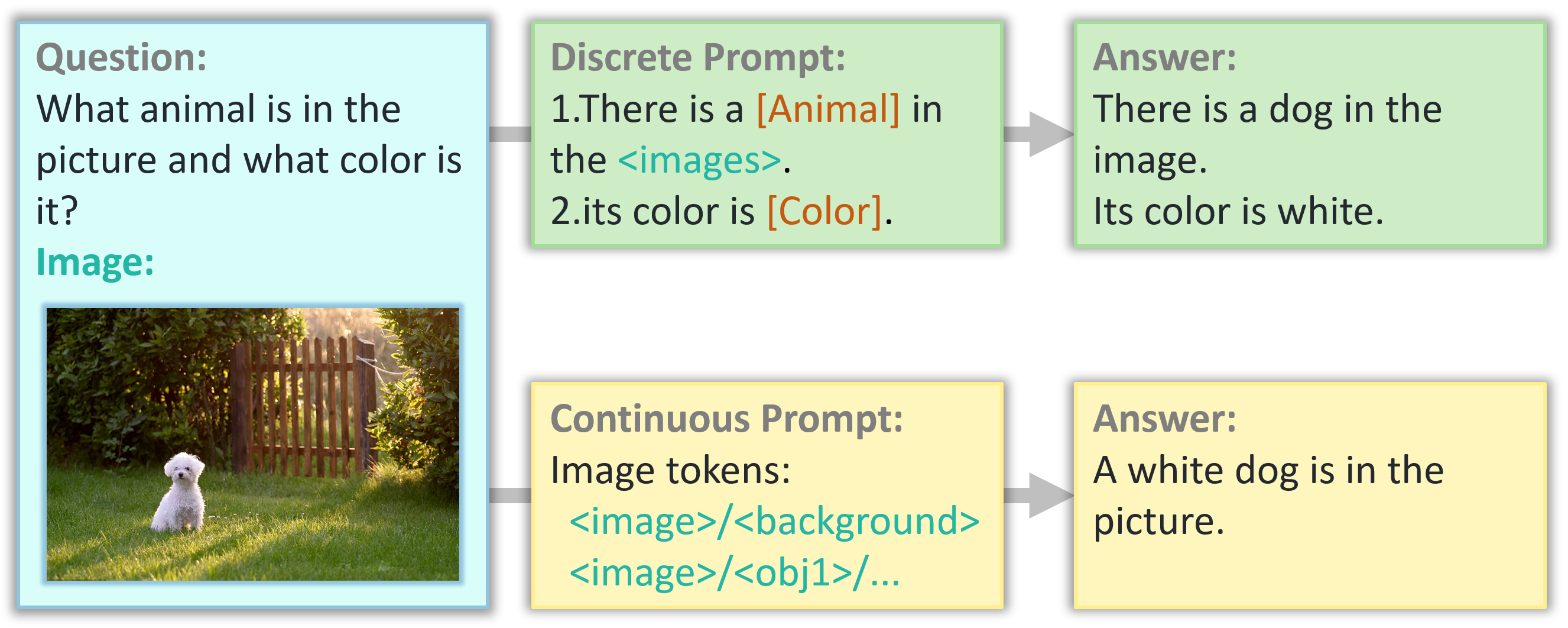}
    \caption{An example showing different prompts and their corresponding responses. It is common to use discrete prompts and continuous prompts in combination, so we will not differentiate between these two methods.}
    \label{fig:p-exp}
\end{figure}

Fig. \ref{fig:p-exp} exemplifies various prompts and their corresponding responses within a Visual Question Answering (VQA) system. 
Upon examination of the figure, it becomes evident that cross-modal discrete prompts amalgamate multimodal data with textual components to create prompts comprising distinct tokens. These tokens, in turn, direct LLMs to generate responses that align with the provided tokens.
In contrast, cross-modal continuous prompts are scrupulously designed to represent multimodal data by utilizing specialized CMR tasks. 

Enhancements to commanding LLMs for CMR often involve prompt tuning \cite{Lester2021ThePO}, a process that optimizes LLMs with task-specific datasets using newly introduced prompt tokens. This progression commences with extracting multimodal data from varied inputs, subsequently, unique prompts are forged from textual input via techniques such as prompt engineering \cite{reynolds2021prompt, Argyle2022AnIA}. Appropriate tuning techniques are implemented to train the models and construct the final answers. In the process of discrete prompts tuning, LLMs are typically kept frozen while semi-freezing, or partial tuning is predominant in the sphere of continuous prompts and instructions tuning. 

CAT \cite{Wang2023CaptionAI} integrates multimodal controls for image captioning by converting visual prompts into mask prompts and using a visual chain-of-thought technique to generate and refine captions based on genre. 

KAT \cite{gui2021kat} combines implicit knowledge extraction from GPT-3  \cite{NEURIPS20201457c0d6} with explicit contrastive learning via CLIP to develop prompts that guide LLMs, focusing on visually-aligned entities. REVIVE \cite{lin2022revive} extends KAT by incorporating regional features to create context-aware prompts, retrieving both implicit and explicit knowledge, and merging visual data for answer generation. 

Visual ChatGPT \cite{wu2023visual} enables users to interact with ChatGPT via text and image inputs, leveraging multiple Visual Foundation Models (VFMs) to handle CMR tasks with a prompt system that details each VFM's function and use case.

PaLM-E \cite{driess2023palm}, derived from PaLM \cite{Chowdhery2022PaLMSL}, demonstrates exceptional performance in a variety of embodied reasoning tasks, including sequential robotic manipulation planning, VQA, and Cross-Modal Captioning (CMC). In PaLM-E, an approach is introduced whereby the Language and Vision model is kept fixed, and only the input encoders are trained to evaluate and compare different-modality encoders. The objective is to establish a strong grounding of the model in observations and enhance its embodiment capabilities. This training methodology can be considered as a type of input-conditioned soft-prompting technique.

VL-T5 \cite{cho2021unifying} is a multimodal model that builds upon T5 \cite{raffel2020exploring}. The model effectively leverages text prefixes to accommodate a wide range of tasks and integrates visual embeddings obtained from a Faster R-CNN \cite{ren2015faster} into a bidirectional multimodal encoder. By utilizing the technique known as Prefix Tuning \cite{li2021prefix}, VL-T5 constructs continuous prompts using diverse prefixes and visual embeddings.

BLIP-2 \cite{li2023blip} presents a two-stage framework designed to mitigate the modality gap in CMR. In the first stage, a querying transformer is pre-trained using a fixed image encoder to facilitate vision-language representation learning. In the second stage, either OPT \cite{zhang2022opt} or FlanT5 \cite{Chung2022ScalingIL} is employed for generative vision-to-language learning. Specifically, FlanT5 is pre-trained using a prefix language modeling loss, where continuous prompts are constructed by combining prefix text and visual representation, while the generation target is represented by the suffix text.

VPGTrans \cite{zhang2023transfer} is a framework that accelerates transfer learning while preserving performance. It leverages a learned mapping between visual content and a soft prompt from a specific LLM. The soft prompt is converted to a different LLM, allowing the converter to merge with the projector trained on another dataset. Word embeddings from both models serve as soft prompts, exhibiting similarity.

eP-ALM \cite{shukor2023ep} is a model that effectively integrates perceptual encoders with language models through the use of continuous prompts, while simultaneously minimizing the number of trainable parameters. This is achieved through the implementation of a singular linear projection layer, which connects the two modules. Leveraging the hierarchical representation encoded in the pre-trained models, eP-ALM tactfully injects [CLS] tokens from multiple layers of the perceptual model \cite{dosovitskiy2020image} into several layers of LLM. Such an approach enables the language model to effectively integrate perceptual input, thus contributing to contextually relevant outputs.

VCoT \cite{Rose2023VisualCO} is a framework that integrates the efficiency, robustness, and multi-step reasoning abilities of Chain of Thought (CoT) with the multimodal capabilities of vision-language models. It aims to resolve logical inconsistencies in sequential datasets by generating synthetic multimodal infillings. To accomplish this, VCoT employs GPT-3 to produce a summary that optimizes the likelihood of the projected output obtained from a multipoint foveation process, augmented by diverse few-shot exemplars.

\subsection{Instruction Tuning}
Instruction tuning is a specialized methodology applied to the training of machine learning models, with a particular emphasis on LLMs. This approach incorporates task-specific directives, designed to enhance the zero-shot performance of these models. 
Cross-modal instruction tuning is an extension of this method, focusing on reinforcing the models' ability to comprehend and respond efficiently to instructions that amalgamate visual or other data modalities. The primary objective of this technique is to bolster the models' capabilities in executing intricate tasks that necessitate CMR, which combines diverse data modalities.

Significantly, instruction tuning presents a notable divergence from traditional prompting systems, primarily due to its emphasis on the necessity of high-quality instruction tuning data during the intensive fine-tuning stage of the models. Nonetheless, it warrants mention that the utility of instruction tuning data is not confined to an individual task but rather spans a range of cognate tasks. Consequently, the acquisition of instruction-following data emerges as a vital constituent in effectuating successful instruction tuning. For instance, MultiInstruct \cite{xu2022multiinstruct} is a comprehensive multimodal instruction tuning benchmark dataset, encompassing 62 diverse tasks across 10 categories in a unified sequence-to-sequence format. Derived from 21 existing open-source datasets, each task is accompanied by five expert-written instructions. utilizing MultiInstruct for instruction tuning, OFA \cite{wang2022ofa}, demonstrated enhanced zero-shot performance when evaluated on the extensive NATURAL INSTRUCTIONS dataset \cite{mishra2021cross}.

InstructBLIP \cite{dai2023instructblip} compiles a broad array of accessible vision-language datasets, converting them into an instruction-tuned format to enhance the diversity of the instruction-tuning data.InstructBLIP advances vision-language instruction tuning and assesses generalizability. It leverages a Q-Former \cite{li2023blip} to extract instruction-sensitive visual features and applies language modeling loss for response synthesis.

To generate instruction-following data that engages visual content, LLaVA \cite{liu2024visual} harnesses the language-only GPT-4 or ChatGPT, characterized as a formidable mentor. More specifically, when encoding an image into its visual features to instigate a textual prompt, LLaVA employs two varieties of symbolic representations: Firstly, Captions, which typically illustrate the visual scenario from multiple vantage points; Secondly, Bounding boxes that generally localize the objects present in the scene, with each box encoding the object concept along with its spatial location. LLaVA-1.5 \cite{liu2024improved} builds upon LLaVA \cite{liu2024visual} by incorporating an improved MLP cross-modal connector and leveraging academic task data to enhance multimodal understanding and data efficiency. It is capable of scaling to high-resolution inputs and achieves state-of-the-art performance with reduced data and a more streamlined architecture.Video-ChatGPT \cite{maaz2023video} elevates video discourse by integrating video representations with an LLM, drawing from vision-language model methodologies for video-related tasks. It repurposes LLaVA \cite{liu2024visual} on prediction tokens in line with its intrinsic autoregressive paradigm.

LLaVA-Med \cite{li2023llava} introduces a curriculum learning approach to adapt LLaVA \cite{liu2024visual} for the biomedical domain, leveraging a self-generated biomedical multimodal instruction-following dataset. Similarly, in the biomedical field, MedVInt \cite{zhang2023pmc} employs a tripartite architecture comprising a visual encoder, language encoder, and multimodal decoder. As a generative learning model, MedVInt aligns a pre-trained vision encoder with a large language model through visual instruction tuning.

GPT4Tools \cite{yang2023gpt4tools} leverages GPT-3.5 to create instruction datasets for tool utilization, enhancing LLMs' ability to perform tool-related tasks. It fine-tunes LLMs using an auto-regressive training paradigm and applies LoRA \cite{hu2021lora} to optimize rank-decomposition factors in the Transformer architecture, preserving the original model parameters. While prefix and suffix prompts are part of the tuning process, they are not discussed here to focus on the fine-tuning method.

MiniGPT-4 \cite{zhu2023minigpt} integrates a static visual encoder with Vicuna \cite{chiang2023vicuna} through a linear projection layer, facilitating the alignment of visual and textual features to optimize learning efficiency. The model undergoes extensive pretraining with aligned image-text pairs, fostering a sophisticated understanding of the interaction between visual and linguistic modalities. Building upon this foundation, MiniGPT-5 \cite{zheng2023minigpt} introduces the concept of generative vokens, allowing training on raw imagery without the need for detailed annotations. It employs a bifurcated training strategy, incorporating classifier-free guidance, a dedicated mapping module, and supervisory losses to ensure precise alignment between tokens and their corresponding image features.

VideoChat \cite{Li2023VideoChatCV} converts videos into textual and embedded formats to enhance multimodal understanding in LLMs. It features a comprehensive video-centric dataset with detailed textual and dialogic annotations. A joint training protocol, leveraging existing image instruction data \cite{zhu2023minigpt, liu2024visual}, improves the system’s spatial perception and reasoning for both static images and dynamic videos.

LaVIN \cite{luo2023cheap}, an efficient multimodal language model employing Mixture-of-Modality Adaptation (MMA), adeptly facilitates CMR tasks with minimal pre-training while preserving strong NLP capabilities. Demonstrating high efficiency and robust performance, LaVIN is ideally suited for chatbot applications and its resources are openly available for scholarly research and development.

Video-LLaMA \cite{zhang2023video} is designed for video-grounded conversations by integrating a language decoder with pre-trained unimodal models for visual and audio inputs. It employs cross-modal pre-training to capture the relationships between multimodal data and is fine-tuned using curated instruction data from various sources \cite{zhu2023minigpt, Li2023VideoChatCV, liu2024visual} to enhance its conversational abilities.

Instruction tuning in ChatBridge \cite{zhao2023chatbridge} enhances the model's multimodal understanding and adherence to human instructions, promoting better zero-shot generalization across various multimodal tasks. This process involves a specialized dataset with multimodal instructions for targeted refinement.

DetGPT \cite{pi2023detgpt} synergizes multimodal models with open-vocabulary object detectors to seamlessly integrate visual and textual modalities. The approach exploits extant corpora dedicated to image captioning and object detection tasks. The enhancement of its interpretative faculties concerning human directives, as well as its competency in generating corresponding object enumerations, is achieved through precision fine-tuning with the aid of a selectively assembled query-answer instruction dataset. The fine-tuning is particularly concentrated on the refinement of the linear projection layer. 

Macaw-LLM \cite{lyu2023macaw}, an integrative model for CMR, incorporates modality, alignment, and cognitive modules. Human-verified instruction-response pairs are generated by GPT-4. The model utilizes a one-step fine-tuning method for instruction adaptation, promoting coherent cross-modality alignment and reducing error transmission.

LLaMA-Adapter \cite{zhang2023llama} refines LLaMA into an instruction-following model using learnable prompts and zero-initialized attention for parameter-efficient fine-tuning \cite{liu2022few}, improving response quality and benchmark performance. LLaMA-Adapter V2 \cite{gao2023llama} further enhances this by refining bias in linear layers, applying joint training with disjoint parameters, and integrating visual data to improve zero-shot multimodal reasoning and task execution.

MiniGPT-v2 \cite{chen2023minigpt} builds vision-language knowledge by using weakly-labeled datasets for diversity and fine-grained datasets for precision. It integrates a frozen ViT \cite{dosovitskiy2020image} with LLaMA-2 \cite{touvron2023llama}, mapping visual tokens to the LLaMA-2 space. Fine-tuning with multi-modal instruction datasets improves conversational abilities, with varying data sampling ratios for fine-grained and new instructions.

PandaGPT \cite{su2023pandagpt} integrates ImageBind's \cite{girdhar2023imagebind} multi-modal encoders with Vicuna's language models to handle vision- and audio-based tasks. It aligns the feature spaces of ImageBind and Vicuna using 160k image-language pairs, optimizing a linear projection matrix and LoRA \cite{hu2021lora} weights while keeping the original parameters unchanged. 

MultiModal-GPT \cite{gong2023multimodal} uses a vision encoder, perceiver resampler, and linguistic decoder, applying LoRA to modify self-attention and cross-attention mechanisms for next-token prediction and sequence generation. 

Ying-VLM \cite{li2023m}, built on BLIP-2 \cite{li2023blip} and Ziya-13B \cite{zhang2022fengshenbang}, is tuned with the M$^3$IT dataset through a dual-phase process, aligning visual features with text via image captioning and instruction tuning for CMR tasks.

Polite Flamingo \cite{chen2023visual} is a method enhancing dataset politeness for vision-language tasks by transforming annotations for instruction tuning. The method involves the filtration and transformation of raw annotations into a more polite style, which subsequently serves as the basis for visual instruction tuning.

NExT-GPT \cite{wu2023next}, designed for high adherence to user directives, yields pertinent multimodal outputs via advanced instruction tuning. Said tuning entails paired training with user instructions and target outputs, applying LoRA \cite{hu2021lora} for precise parameter updates. Reinforced by optimization using premium annotations, the model incorporates decoding-level fine-tuning, aligning output tokens with the intended multimodal context.

ChatSpot \cite{zhao2023chatspot}, a multimodal interaction model, achieves language-image alignment through a streamlined MLP approach, foregoing additional models or post-processing. This facilitates granular user interactions with image regions, enhancing region-specific instruction comprehension through precise tuning. Trained on the MGVLID, it demonstrates improved task performance.

BLIVA \cite{hu2023bliva} leverages a Q-Former \cite{li2023blip} and a fixed image encoder for instruction-tailored visual feature extraction. The resultant features serve as inputs for an LLM which, following pre-training with image-caption pairs, aligns itself with visual information. This pre-trained LLM is then capable of generating descriptive image representations.

BuboGPT \cite{zhao2023bubogpt} employs visual grounding for enhanced CMR, using a two-stage training approach with a curated instruction dataset to improve multimodal comprehension. The first stage synchronizes the linear projection layer output with the lexical embedding space, while the second conditions the LLM to interpret directives and produce modality-specific responses.

VisionLLM \cite{wang2023visionllm} features a unified linguistic directive, an image tokenizer, and an LLM-driven task decoder, facilitating the performance of assorted tasks steered by textual prompts. This consolidated instruction set is adept at handling both standalone vision and composite vision-language tasks, enhancing task adaptability. 

LAMM \cite{yin2023lamm} constitutes a scalable architecture for augmenting multi-modal LLMs (MLLMs) with extra modalities such as point clouds. It introduces the LAMM-Dataset, purposed for detailed vision tasks, and establishes the LAMM-Benchmark, the inaugural comprehensive standard for assessing MLLMs, covering instruction tuning and model expansion proficiencies.

Qwen-VL \cite{bai2023qwen}, a suite of expansive vision-language models, is crafted to interpret text and imagery. Following its pre-training, Qwen-VL is subjected to supervised fine-tuning, augmenting its directive adherence and conversational abilities via multi-modal instruction tuning. 

mPLUG-Owl \cite{ye2023mplug} employs joint instruction tuning via LoRA, aligning uni-modal and multi-modal data to enhance response quality without requiring separate vision-language realignment. Training focuses on language modeling, optimizing token prediction, and improving multi-modal performance. mPLUG-Owl2 \cite{ye2023mplug2} refines instruction tuning and modality adaptation, advancing vision encoding and text processing for stronger integration of visual and linguistic information. mPLUG-Owl3 \cite{ye2024mplug} further improves efficiency in handling diverse visual inputs and introduces the Hyper Attention module to enhance cross-attention and integration between modalities.

X-LLM \cite{chen2023x} is a framework that seamlessly combines pre-trained single-modal encoders with LLMs using multiple interfaces. Specifically, in the context of image processing, X-LLM employs a Q-Former and an I-Adapter module. The Q-Former \cite{li2023blip} component is responsible for transforming image features into quasi-linguistic embeddings. Meanwhile, the I-Adapter module ensures the alignment of these embeddings with the LLM's embedding dimension, enabling effective cross-modal reasoning.

LION \cite{chen2024lion} improves CMR by overcoming the constraints of pre-training through the utilization of rough image-text pairs. It incorporates spatial-aware visual information and employs a three-step tuning approach to enhance comprehension of visual and textual content. Similarly, VistaLLM \cite{pramanick2024jack} emphasizes fine-tuning at different levels of detail, combining tasks of varying granularity using multiple images.

Based on BLIP-2 \cite{li2023blip}, xGen-MM (BLIP-3) \cite{xue2024xgen} introduces instruction tuning to enhance multimodal model capabilities. It leverages richer and more diverse training data and employs a scalable vision token sampler in place of Q-Former layers. Additionally, it unifies training objectives with a single loss function across all stages. These improvements enhance performance on visual language tasks and deliver competitive results across various benchmarks.

\subsection{Multimodal Pre-training Foundation Models}

Multimodal pre-training is a methodology used to train models to acquire knowledge from various data sources such as images, text, and audio. The goal is to develop a comprehensive understanding of the world by leveraging extensive datasets. These models effectively grasp the complex interplay between different modalities and generate rich representations that capture key semantic information.

Flamingo \cite{alayrac2022flamingo}, a cutting-edge Visual Language Model (VLM), adeptly amalgamates textual and visual information to swiftly adapt to cross-modal reasoning tasks with limited data. Utilizing pre-trained models, the Perceiver Resampler converts features from the Vision Encoder into tokens, conditioning the language model through cross-attention layers. This innovative approach enables accurate prediction of subsequent tokens, generating coherent, contextually relevant responses. Throughout the training process, Flamingo selectively fine-tunes the Perceiver Resampler and language modules, while keeping the Vision Encoder parameters fixed.

Cheetor \cite{li2023empowering} endeavors to leverage the Visual Prompt Generator (VPG) for improved performance in CMR tasks, with a particular focus on VQA. Currently, available VPG approaches, such as linear projection, Resampler, and Q-former, extract visual prompts from densely encoded image features obtained through vision backbones, exemplified by ViT \cite{dosovitskiy2020image}.

MAGMA \cite{eichenberg2021magma}, by converting image features into language embeddings, adapts language transformers to visual data without complete retraining. It comprises four elements: a Visual Encoder that generates feature vectors from images, an Image Prefix module that translates these vectors into embeddings, an autoregressive Language Model integrating these embeddings, and Adapter layers tuned in situ within the transformer language model. Throughout the training, the Language Model's architecture remains static, utilizing pre-existing GPT-J weights \cite{wang2021gpt}. Simultaneously, the Visual Encoder draws on pre-trained CLIP weights, with the Image Prefix and Adapters undergoing training from the ground up.

Prismer \cite{liu2023prismer}, an efficient encoder-decoder model, leverages a pre-trained expert library for handling multimodal inputs and generating text. It employs a vision encoder for RGB images and labels, alongside an autoregressive text decoder. Key trainable elements—the Experts Resampler and the Adaptor—normalize input variability and enhance vision-language processing. Pre-trained weights remain largely fixed to preserve domain knowledge, optimizing Prismer for tasks like image captioning and visual question answering. Training focuses on sequentially predicting text tokens.

PaLI \cite{chen2022pali} is a multimodal model that employs the ViT-e \cite{zhai2022scaling} module for image processing, and the mT5 \cite{xue2020mt5} module for language processing. It outperforms ViT-G in vision-language tasks - an indication that larger ViT \cite{dosovitskiy2020image} backbones could yield even better results. Training PaLI involves various tasks, including pure language understanding tasks, to maintain mT5's language competencies without forgetting. Notably, only the language component gets updated during training as the vision component remains static.

Utilizing a dual-encoder architecture like CLIP \cite{radford2021learning} with a Visual Encoder based on TimeSformer (TSF) \cite{bertasius2021space}, LAVILA \cite{zhao2023learning} generates textual descriptions for video clips through two LLMs, NARRATOR and REPHRASER. NARRATOR pseudo labels existing and new clips with narrations, while REPHRASER paraphrases existing narrations. The combined output of these LLMs trains the dual encoders, with the NARRATOR's Video Encoder remaining static.

VideoCoCa \cite{yan2023videococa}, leverages the pre-trained CoCa \cite{yu2022coca} by maintaining all of its weights and avoiding the need for additional module learning. Instead, VideoCoCa calculates offline frame token embeddings from the already frozen CoCa image encoder. These tokens then undergo processing by both a generative and a contrastive pooler to produce an effective zero-shot video-text baseline. The model continues to pre-train on video-text data and optimizes attentional poolers and text decoders using contrastive loss and captioning loss. By keeping the image encoder static, the model minimizes the computational burden on frame embedding.

Inspired by Flamingo \cite{alayrac2022flamingo}, OpenFlamingo \cite{awadalla2023openflamingo} is a framework that generates autoregressive vision-language models utilizing CLIP for image encoding and open-source language models for decoding. It handles interleaved sequences of images and text tokens, facilitating their interaction through dense cross-attention modules. Furthermore, the framework embeds images utilizing a trainable Perceiver resampler. OpenFlamingo undergoes training using a combination of image-text pairs and interleaved image-text sequences. 

Otter \cite{li2023mimic}, an in-context instruction-tuned multimodal model, derives from OpenFlamingo \cite{awadalla2023openflamingo}, a large-scale, interleaved image-text pretraining foundation. It integrates linguistic and visual inputs to augment perception, reasoning, and planning functions. Designed for contextual instruction adherence, Otter excels in multi-turn conversational reasoning and acts as an egocentric visual assistant, leveraging video and image sequences for indoor navigation and event planning.

Gato \cite{ reed2022generalist} is designed to train on diverse data types such as images, text, observations, and actions. The information is then converted into sequential tokens to effectively handle this multi-modal data. During deployment, Gato utilizes these tokens to generate contextual responses or actions. By employing a joint visual-language model, Gato is capable of understanding and generating responses using both text and visual data, enabling it to handle tasks that involve the comprehension of both visual and textual information.

The enhanced PaLI-X \cite{chen2023pali} model represents an advancement of the original PaLI, emphasizing Optical Character Recognition (OCR) capabilities with a substantial visual component. This model utilizes a more complex, two-stage training process wherein only the augmented visual encoder \cite{dehghani2023scaling} remains unchanged in the initial stage. In the subsequent stage, potential augmentations to the visual component correspond with model progression and escalated image resolution parameters. The PaLI-X model employs an extensive UL2 \cite{tay2022ul2} encoder-decoder infrastructure comprising 50 equivalent layers and effectively integrates visual and token embeddings.

The architecture of COSA \cite{chen2023cosa} combines visual and textual information using a vision encoder based on ViT \cite{dosovitskiy2020image} and a text encoder based on BERT. It introduces cross-attention layers between self-attention and feed-forward layers of BERT to enable interaction between modalities during the forward pass. This design allows the text encoder to serve as a single-modality encoder, cross-modality encoder, and decoder, making COSA versatile for processing and interpreting information from different modalities and generating outputs.

Like Flamingo, GIT \cite{wang2022git} consolidates vision-language tasks with streamlined architecture and increased pre-training data for enhanced performance. However, in GIT, all parameters are updated to optimize fitting for vision-language tasks. It employs a single language modeling task removes external modules, and facilitates the exploration of design options. The model undergoes 2 epochs of training, operating as a decoder-only language model, with the potential for refinement through text-only data utilization.

BEIT-3 \cite{wang2022image} is a multimodal foundation model that undergoes masked data modeling during pretraining on both monomodal and multimodal data. It employs a shared Multiway Transformer network \cite{bao2022vlmo} with modality-specific experts to encode diverse modalities. This approach allows for the capture of modality-specific information and facilitates alignment between different modalities through a shared self-attention module. Consequently, BEIT-3 proves advantageous in performing a range of vision and vision-language tasks.

MetaLM \cite{hao2022language} is a versatile language model crafted to interface with an array of foundation models, integrating pre-trained encoders for multiple modalities and a language model for general task-based free-text generation. This hybrid model amalgamates the strengths of language and foundation models, undergoing fine-tuning through task-specific instructions. Its zero-shot generalization capacity is assessed by omitting analogous datasets in the testing phase.

KOSMOS-1 \cite{huang2023language} is a multimodal language model based on a Transformer-based causal language model, trained on multimodal corpora for zero-shot and few-shot evaluation on both language and multimodal tasks. KOSMOS-2 \cite{peng2023kosmos} surpasses KOSMOS-1, offering improved referential and grounding abilities by processing user-specified image regions and producing linked visual-textual outputs. Training incorporates grounded pairs, continuous bounding box-to-location token conversion, and a hyperlink format to bridge image regions with text.

Unified-IO \cite{lu2022unified}, a T5-derived \cite{raffel2020exploring} transformer model, employs an encoder-decoder framework with stacked self-attention and cross-attention layers, complemented by feed-forward networks. Pre-trained with unsupervised data from text, imagery, and paired formats, it subsequently progresses through extensive multi-task training on a variety of tasks. Yet, it forgoes task-specific fine-tuning, a strategy suggested by research to potentially enhance performance.

mPLUG-2 \cite{xu2023mplug} cutting-edge development in multi-modal pre-training foundation models, addressing multi-modal interaction and entanglement challenges. It features a strategic module-based design that enhances cross-modality collaboration with shared modules, while modality-specific modules resolve entanglement issues. Jointly trained on diverse datasets, mPLUG-2 supports a broad spectrum of downstream tasks across text, image, and video modalities.

MaMMUT \cite{kuo2023mammut} is a foundational multimodal pre-training model that integrates a single image encoder and text decoder, effectively combining contrastive learning with autoregressive prediction. This design enables seamless handling of vision and language tasks, supports direct adaptation to video, and facilitates task extension. Its two-pass learning approach, utilizing shared model weights, ensures efficient training and inference across various modalities.

\section{LLM as Textual Processor}

\begin{figure}
    \centering
    \includegraphics[width=0.8\linewidth]{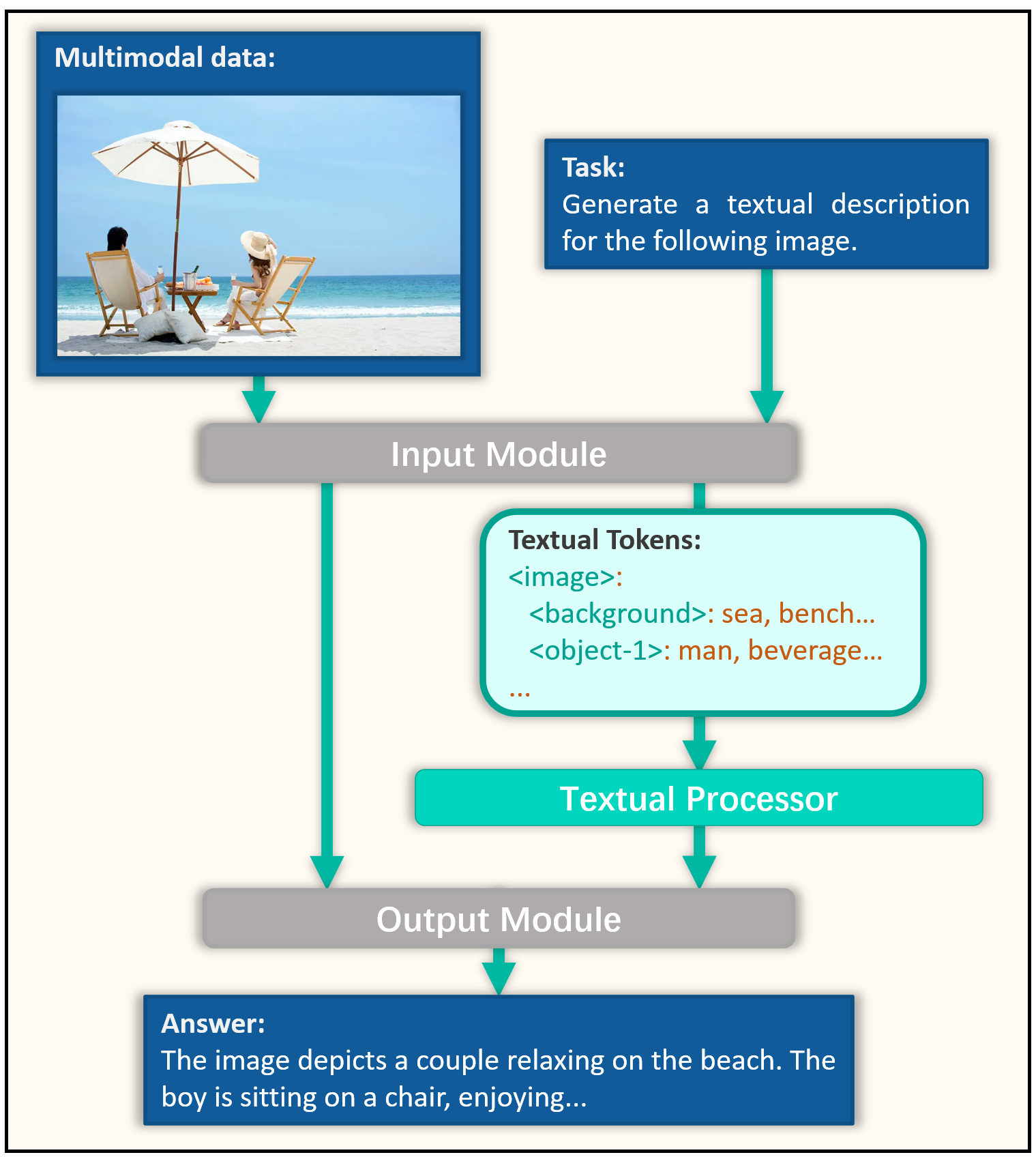}
    \caption{The function of Large Language Models (LLMs) as Textual Processor within cross-modal reasoning frameworks.}
    \label{fig:exp-TP}
\end{figure}
Figure \ref{fig:exp-TP} demonstrates the role of Large Language Models (LLMs) as Textual Processors, optimizing linguistic content for task-specific applications. Certain modules generate textual input for LLMs, which subsequently process this input to produce output for downstream modules. LLMs generate specific textual content to inform model design, with the resulting output potentially serving as prompts or responses. LLMs demonstrate particular efficacy in text processing, efficiently filtering extensive corpora to extract task-relevant elements and synthesizing them into targeted responses.

Based on distinct text processing methodologies, we propose a bifurcation of these models into two categories: semantic refiner and content amplifier.

\subsection{Semantic Refiner}

Socratic Models \cite{zeng2022socratic} utilize multimodal prompting to integrate information from non-language domains into language prompts, enabling language models to reason using multimodal information. This is achieved by substituting entities described in language prompts with entities from other modalities within discrete prompts, facilitating the integration of knowledge across different modalities within the SM framework.

PromptCap \cite{hu2022promptcap} is a cross-modal reasoning model that integrates natural language prompts for image captioning. It takes an image-caption sample and a natural language prompt as inputs and is specifically trained to generate captions that aid LLMs in question answering.

Img2LLM \cite{guo2022images} is a module designed to enable LLMs to engage in VQA. The core concept behind Img2LLM involves leveraging a combination of vision-language models and question generation models to convert image content into Exemplar Prompts, comprising synthetic question-answer pairs. Additionally, Caption Prompts are generated from textual modalities. As illustrated in Figure 1, prompts serve as a bridge between diverse modalities and tasks within the LLM, allowing for effective VQA execution without the necessity of task-specific training.

MoqaGPT \cite{zhang2023moqagpt} is also a framework that leverages different models to extract answers from diverse modalities and employs LLMs for CMR tasks. It functions in a zero-shot fashion, supporting multiple modalities without necessitating joint representation or inference. The responses are derived from retrieved results, guaranteeing reliability, and all intermediate outputs are expressed in natural language, enhancing transparency and interoperability.

VAST \cite{chen2023vast} represents an approach to multimodal integration, incorporating tripartite encoding mechanisms designated for visual, auditory, and textual subtitle modalities. Its architecture privileges a text-based encoder tasked with executing cross-modality fusion by employing cross-attention mechanisms. VAST's training regimen is conducted on a richly diverse dataset of omni-modal video captions, equipping the model with the faculty to perceive and interpret a quartet of modalities. 

\subsection{Content Amplifier}

ViewRefer \cite{guo2023viewrefer} is a multi-view framework designed for 3D visual grounding. It leverages the capabilities of LLMs to enrich the input grounding text with descriptions related to different views. By incorporating view cues from both textual and 3D modalities, ViewRefer enhances the understanding of spatial inter-object relations. The utilization of expanded texts generated by LLMs allows for the integration of multi-view semantics, effectively leveraging the linguistic knowledge of LLMs to attain improved grounding performance for target objects.

PointCLIP V2 \cite{zhu2023pointclip} is a framework that seamlessly integrates both CLIP \cite{Radford2021LearningTV} and GPT-3, enabling the achievement of comprehensive and unified 3D open-world understanding. By leveraging GPT-3 and providing it with 3D commands as prompts, PointCLIP V2 generates highly descriptive 3D semantic text, leading to enhanced alignment between images and text. Notably, this approach significantly improves 3D understanding without the requirement of extensive 3D training data.

ScienceQA \cite{lu2022learn}, a benchmark for science question answering that includes a broad range of multimodal multiple-choice questions. Utilizing CoT, they adapt discrete prompts to guide LLMs, including GPT-3 \cite{NEURIPS20201457c0d6} and UnifiedQA \cite{Khashabi2020UnifiedQACF}, to generate answers that accurately reflect the reasoning process when answering ScienceQA questions. By leveraging CoT, LLMs can generate meaningful explanations that showcase their reasoning ability when answering complex questions.

IdealGPT \cite{You2023IdealGPTID} adopts an iterative and multi-round approach to address CMR tasks. Its framework incorporates manually designed diverse prompts that fulfill distinctive roles within LLMs, namely Questioner, Answerer, and Reasoner. By iteratively decomposing vision and language reasoning utilizing LLMs, IdealGPT effectively tackles the CMR task.

\section{LLM as Cognitive Controller}

\begin{figure}
    \centering
    \includegraphics[width=0.8\linewidth]{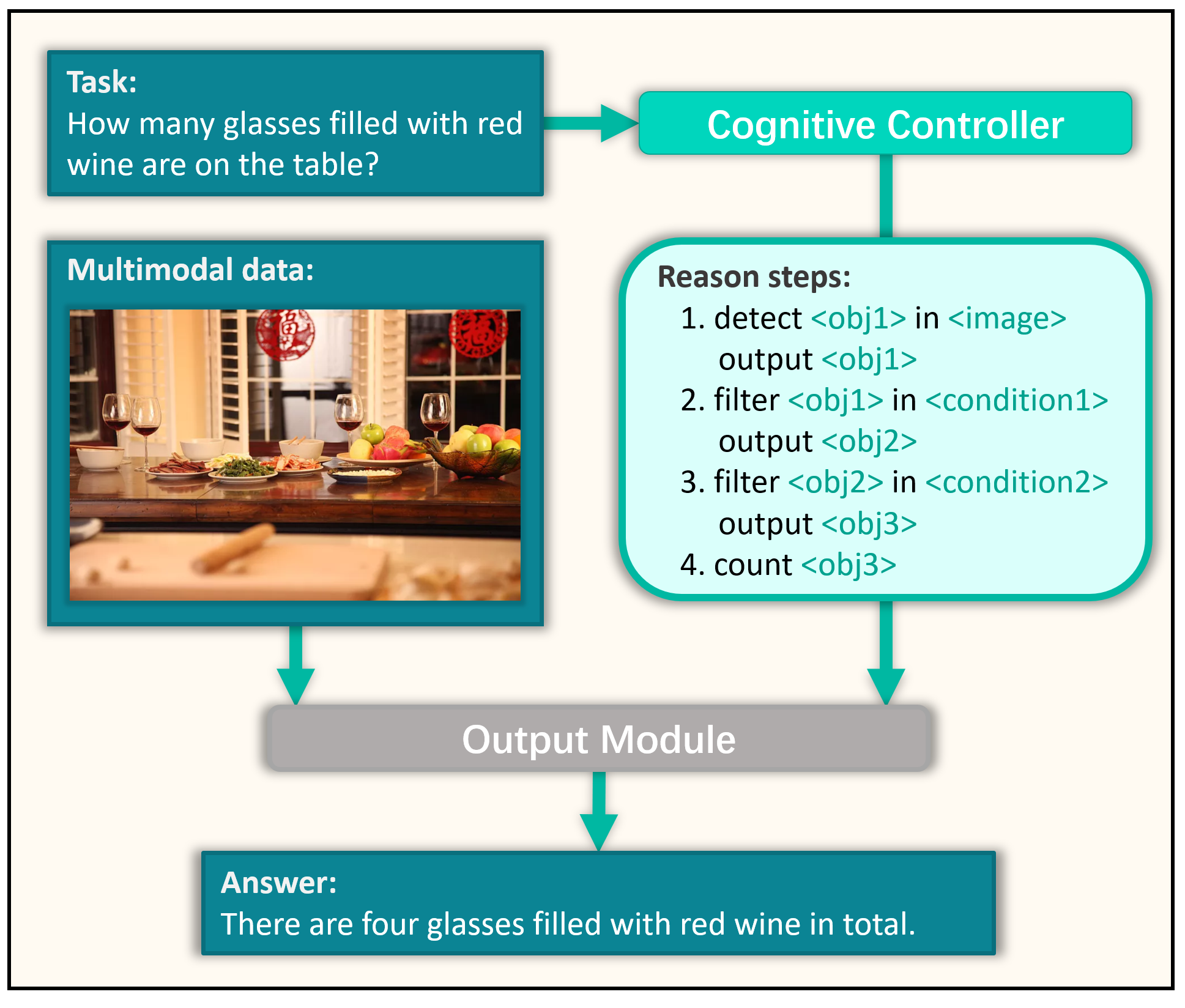}
    \caption{The function of Large Language Models (LLMs) as Cognitive Controller within cross-modal reasoning frameworks. }
    \label{fig:exp-CC}
\end{figure}

Fig. \ref{fig:exp-CC} depicts the role of Large Language Models (LLMs) in facilitating the architecture of the reasoning workflow. Within this framework, tasks and the corresponding multimodal data are inputted into the LLMs. These models are capable of constructing a reasoning blueprint that outlines the entire procedure. The execution of this process is delegated to auxiliary modules. The reasoning workflow is compartmentalized into discrete phases, with each step's inputs and outputs being meticulously architected by the LLMs. In addition, the LLMs possess the capability to determine the appropriate tools or models that should be employed at various junctures of the reasoning process.

\subsection{Programmatic Construction}

AssistGPT \cite{gao2023assistgpt} is a multi-modal AI framework designed for CMR. It employs a language and code interface spanning four core modules: Planner, Executor, Inspector, and Learner, which collectively orchestrate reasoning, facilitate tool execution, monitor inputs, and assess performance. Central to this system, the GPT-4-driven Planner module synthesizes various inputs—ranging from instruction prompts and tool set visuals to context-specific examples—and constructs subsequent prompts to advance the reasoning process.

ProgPrompt \cite{singh2023progprompt} enhances LLMs with a Cognitive Controller for Python programs, imported actions, and environment objects, significantly augmenting their capabilities for CMR tasks. By capitalizing on the inherent training of LLMs in programming languages, ProgPrompt streamlines the prompt construction process by seamlessly integrating pertinent environmental information and necessary primitive actions for successful task execution. ProgPrompt provides LLMs with discrete prompts in the form of sample tasks and plans, enabling them to better grasp the context and generate more accurate and appropriate responses. 

Inspired by ProgPrompt, VisProg \cite{gupta2023visual} is a system specifically developed to tackle CMR tasks. Its architecture incorporates the utilization of in-context learning, utilizing the power of GPT-3, to dynamically generate programs based on novel instructions. These programs are subsequently executed on input images, enabling the system to accurately generate predictions. A key advantage of this approach is the ability to employ discrete prompts, such as decomposer prompts \cite{khot2022decomposed}, enabling the generation of sequential sub-tasks that can be efficiently managed by dedicated sub-task handlers. Through this innovative methodology, VisProg demonstrates great potential for various applications requiring cross-modal reasoning and offers a highly adaptable and versatile solution in the field.

Instruct2Act \cite{huang2023instruct2act} framework adopts a programmatic methodology to facilitate cross-modal reasoning, which is instrumental in reinforcing robotic control mechanisms. It encompasses a suite of Application Programming Interfaces, demonstrative usage cases, and comprehensive task delineations. LLM is proficient at generating Python functions capable of executing specified robotic actions. The methodology presents a plethora of merits, such as enhanced adaptability, obviation of the need for training datasets, and augmented action precision attributable to the foundational model. 

ViperGPT \cite{suris2023vipergpt} is a framework that efficiently combines vision-and-language models into subroutines, akin to the VisProg \cite{gupta2023visual}. Its prompt embodies flexibility and versatility, offering an API for integrating diverse perceptual and knowledge modules. These modules encompass functionalities like object detection, depth estimation, and language model queries. Leveraging this API, ViperGPT facilitates seamless access to these modules, generating precise Python code that produces accurate and contextually relevant results for specified queries.

In a similar way, ProViQ \cite{choudhury2023zero} leverages pre-trained models for visual tasks, incorporating task-specific modules without additional training. Its modular integration enhances capabilities while maintaining interpretable reasoning. An LLM auto-generates Python code based on input questions, visual APIs, and examples, processing video input.

\subsection{Linguistic Interaction}

Within the domain of cross-modal reasoning incorporating large language models, linguistic interaction frequently serves as a pivotal interface among heterogeneous language model architectures. This necessitates the prerequisite that the antecedent module is capable of producing outputs in natural language, which can, in turn, be interpreted by the subsequent module \cite{gao2024cantor}. LLMs are versatile and can be integrated into various components of these sophisticated model frameworks. For example, the previously referenced MoqaGPT \cite{zhang2023moqagpt} embeds LLMs within distinct inferential procedures. Additionally, LLMs are equipped to engage in iterative dialogues or conduct internal discourse, facilitating a process of deep and comprehensive reasoning.

HuggingGPT \cite{shen2023hugginggpt} is an extensive system that integrates with both LLMs and the machine learning community to facilitate CMR. This system effectively employs model descriptions retrieved from the Hugging Face library, which are incorporated as discrete prompts. By utilizing these prompts, ChatGPT is capable of selecting relevant models and generating precise and impactful responses to user queries. 

Similar to HuggingGPT, Chameleon \cite{lu2023chameleon} is an AI system that enhances LLMs by integrating modular components. This system leverages extensive in-context examples to acquire knowledge and generate answers for cross-modal reasoning tasks. Chameleon combines various components, including LLMs, vision models, and web search engines to achieve its objectives. It utilizes discrete prompts to generate precise paradigm sentences that can be easily understood by auxiliary tools.

ChatCaptioner \cite{Zhu2023ChatGPTAB} is an automated questioning system designed specifically for image captioning. It leverages the capabilities of ChatGPT to generate insightful and informative questions that are then directed toward  BLIP-2 \cite{li2023blip}. By assimilating fresh visual information from the answers provided by BLIP-2, ChatCaptioner produces augmented and more detailed image descriptions. The system employs a robust prompting mechanism that incorporates task instructions, chat logs, and question instructions to ensure the generation of contextually aware and relevant questions. 

Similar to ChatCaptioner \cite{Zhu2023ChatGPTAB}, Video ChatCaptioner \cite{chen2023video} employs ChatGPT for inquiry generation and BLIP-2 for response to decipher the video's spatiotemporal features. A stratified prompt hierarchy governs this mechanism, promoting varied, insightful queries and dependable responses.

Shikra \cite{chen2023shikra} is a model that tackles the inherent limitation of natural referential ability. Shikra possesses the capability to process spatial coordinates expressed in natural language, thereby eliminating the need for supplementary components such as position encoders or external models. Moreover, Shikra adopts an uncomplicated and transparent approach to constructing task templates for various tasks.

Inner Monologue \cite{huang2022inner} enables the achievement of complex tasks in simulated and real-world robotics, devoid of supplemental training. Each domain within Inner Monologue employs unique prompts and feedback models. Leveraging an LLM, it uses a series of discrete prompts to interpret human instructions and transforms them into actionable directives for the agent.

MM-ReAct \cite{yang2023mm} integrates ChatGPT and a team of vision experts to enhance CMR. With a tailored textual prompt, it efficiently manages dense visual signals, bolstering ChatGPT's visual comprehension and delivering comprehensive responses. This is achieved through collaborative efforts and an automated selection process.

SayCan \cite{ahn2022can} is a model that enhances CMR tasks, particularly in robotic applications, by extracting and utilizing the knowledge from LLMs. This model consists of two crucial components – the LLM (also known as "Say") and the learned affordance functions (also known as "Can"). To segment a CMR task into distinct steps and guide LLMs, the Say component employs CoT, resulting in clear and succinct prompts.

WALL-E \cite{wang2023wall} is a model that integrates ChatGPT, visual grounding systems, and robotic grasping systems, enhancing human-robot interaction. It utilizes a range of prompts such as explanations, guidance, environmental information, task rules, open vocabulary communication, command confirmation, and execution feedback to facilitate ChatGPT's learning and foster efficient communication between humans and robots. WALL-E offers numerous discrete prompts for LLMs to construct CoT and select objects.

\section{LLM as Knowledge Enhancer}
\begin{figure}
    \centering
    \includegraphics[width=0.8\linewidth]{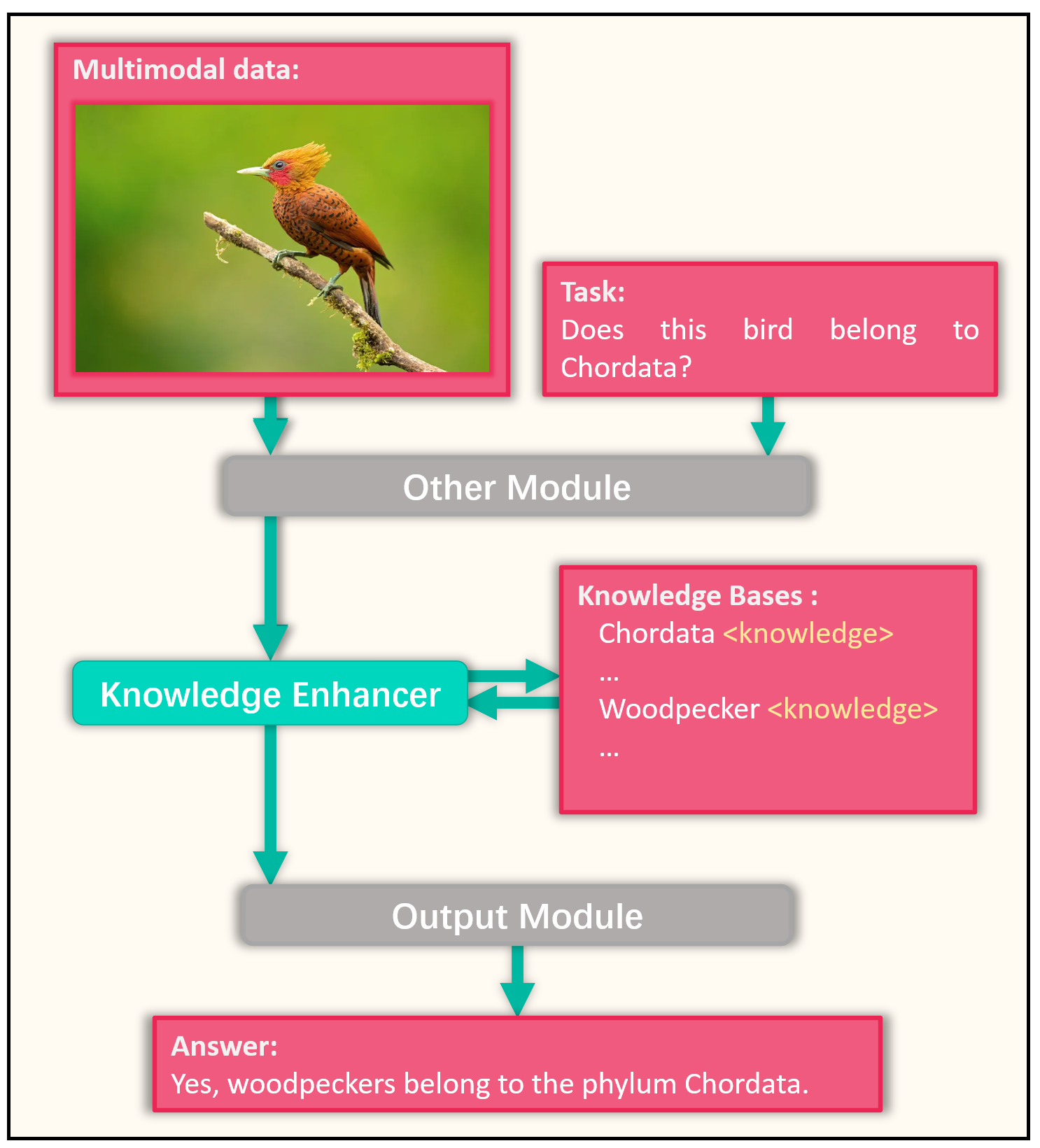}
    \caption{The function of Large Language Models (LLMs) as Knowledge Enhancer within cross-modal reasoning frameworks.}
    \label{fig:exp-ke}
\end{figure}
Figure \ref{fig:exp-ke} depicts how LLMs utilize knowledge bases for completing CMR tasks. Certain models rely on implicit knowledge within LLMs. Furthermore, to cater to specific domains, LLMs are equipped to interact with external databases through various approaches, thereby improving their task performance by leveraging extensive expertise.

\subsection{Implicit Cognition}

Leveraging the implicit cognitive capabilities of Large Language Models (LLMs) is an efficacious approach to enhancing Cross-Modal Reasoning (CMR) tasks. These methodologies employ LLMs to generate nuanced task descriptions, extend the range of potential response options, and integrate common sense into the reasoning process. For illustration, PICa \cite{yang2022empirical} employs a strategy of directly prompting GPT-3 to synchronously assimilate and apply pertinent knowledge for informed reasoning. Similarly, ViewRefer \cite{guo2023viewrefer} utilizes the linguistic adeptness inherent in LLMs to provide rich, multi-perspective descriptions. Concurrently, PromptCap \cite{hu2022promptcap} capitalizes on the expansive knowledge repository contained within LLMs to create complex image captions that have been specifically crafted to assist in resolving targeted visual queries. KAT \cite{gui2021kat} and REVIVE \cite{lin2022revive} harness the power of LLMs to imbue systems with implicit knowledge, which is then synergistically melded with external knowledge sourced from adjunct modules to facilitate enhanced reasoning.

Given that the aforementioned models have been delineated previously, a detailed reexamination will be omitted in this discourse. Instead, the subsequent discourse will advance with a discussion on selected quintessential approaches.

CaFo \cite{zhang2023prompt}, or Cascade of Foundation models, enhances few-shot learning by integrating diverse prior knowledge from different pretraining paradigms. It combines the language-contrastive knowledge of CLIP \cite{Radford2021LearningTV}, vision-contrastive knowledge of DINO \cite{caron2021emerging}, vision-generative knowledge of DALL-E \cite{ramesh2021zero}, and language-generative knowledge of GPT-3 \cite{NEURIPS20201457c0d6}. By using templates to generate prompts rich in linguistic semantics for CLIP's textual encoder, CaFo effectively improves few-shot learning through the integration of these varied knowledge sources.

CuPL \cite{pratt2023does} also employs LLMs for the generation of prompts. In the context of image prompts, CuPL utilizes a series of LLM prompt templates. Although this process entails manual engineering to some extent, it is considerably less time-consuming compared to the conventional method used to create hand-engineered sentences for CLIP's image-prompt templates.

Prophet \cite{shao2023prompting} is a framework developed to enhance the performance of GPT-3 in knowledge-based VQA tasks. The framework incorporates answer heuristics, which represent potential answers that are strategically presented in the prompt, to guide the response generation process. Specifically, a prompt in Prophet consists of a prompt head, a set of in-context examples, and a testing input.

Proofread \cite{zhou2023prompting} is a knowledge-driven framework that integrates vision and language models to enhance the accuracy of VQA. It harnesses the power of an LLM to retrieve explicit knowledge while leveraging a vision language model to provide visual-based answers. To ensure the reliability of the knowledge employed, a knowledge perceiver is integrated into the framework to filter out potentially detrimental information. The framework comprises a dedicated Question Model and Knowledge Answer Model, which collectively generate pertinent questions and corresponding answers. Guided by prompts, knowledge, and questions are meticulously inputted into the models, facilitating the seamless integration of multimodal reasoning.

\subsection{Augmented Knowledge}

Various methodologies have been developed to harness real-time information by interfacing with web search engines \cite{nakano2021webgpt}, import domain-specific knowledge from specialized repositories to establish robust contextual bases \cite{yu2022generate}, and accumulate data through the analysis of GitHub repositories \cite{liu2023ml}. These approaches enhance LLM capabilities by facilitating the acquisition of external knowledge, yet they have not been traditionally exploited for cross-modal tasks. 

To readily incorporate external knowledge into LLMs, it is expedient to utilize ancillary tools. MM-ReAct \cite{yang2023mm} exemplifies such integration by merging LLMs with search engines. This methodology is extensible, allowing other modalities like Bing search to be appended to the foundational model. Furthermore, HuggingGPT \cite{shen2023hugginggpt} bridges the gap between ChatGPT-style LLMs and the Machine Learning (ML) ecosystem, wherein LLMs are tasked with matching specific models to corresponding tasks as per the expert-curated model descriptions available on the Hugging Face platform. Moreover, the Chameleon framework \cite{lu2023chameleon} amalgamates a wealth of tools, encompassing LLMs, pre-trained vision models, web search engines, Python executables, and rules-based systems, to craft an AI construct proficient in CMR tasks.

Beyond the aforementioned strategies, the process of instruction tuning presents another avenue for embedding external knowledge within LLMs. Approaches such as LLaVA-Med \cite{li2023llava} and MedVInt \cite{zhang2023pmc} incorporate domain-specific knowledge through instruction tuning, equipping the models to adhere to the stringent and nuanced demands of the biomedical domain.

\section{Challenges and Future Directions}

In CMR, there is significant potential for expanding the utility of LLMs, offering promising opportunities for future exploration. While existing applications demonstrate the versatility of LLMs in tasks such as image captioning, video understanding, and speech dialog, some substantial limitations have been identified. 
Therefore, in this section, we discuss the current challenges and future directions in the area of CMR with LLMs.

\subsection{Modality Scalability of LLMs}

Currently, CMR with LLMs focuses on integrating and interpreting primarily textual, visual, audio, and video modalities. However, there is a pressing need to expand its application domain to encompass a broader spectrum of modalities. Exploring the scalability of these models presents an efficient pathway to extend their applications across diverse industries, including healthcare, transportation, and beyond.

Recent research has demonstrated promising results in integrating sensor data for cross-modal tasks \cite{xiang2023multi,huang2022multi}. Nevertheless, challenges persist in this integration process. Firstly, the information fusion methodologies can be further refined through more sophisticated approaches. Secondly, current methods, such as MSMDFusion \cite{  jiao2023msmdfusion}, Voxel Field Fusion \cite{li2022voxel}, typically incorporate visual information from a limited set of images, resulting in a paucity of rich visual context. Leveraging LLMs and fusing more comprehensive visual information or multi-modal data streams could significantly enhance performance in these tasks.

Furthermore, the incorporation of additional modalities into CMR tasks holds considerable potential. For instance, integrating biometric and haptic data could open new avenues for application. Biometric data could play a crucial role in advancing medical applications, while haptic data is essential for robotic systems. This expansion of modalities not only broadens the scope of CMR tasks but also enhances their relevance and efficacy across various domains.

\subsection{Hallucination in CMR with LLMs}

Enhancing the reliability of LLMs is crucial for improving the precision of CMR. However, the reliability of safety mechanisms in LLMs remains uncertain because of hallucinations \cite{zhao2023survey}. These models are susceptible to generating hallucinated content—outputs that appear credible but lack factual accuracy \cite{bang2023multitask, li2023evaluating}. This phenomenon significantly compromises the reliability and security of LLM outputs, consequently impacting the integrity of CMR tasks.

One strategy to mitigate the hallucinations is the implementation of Reinforcement Learning from Human Feedback (RLHF) \cite{glaese2022improving}. RLHF integrates human feedback into the training process, aiming to align LLMs more closely with ethical and factual standards. Furthermore, there is a critical interdependence between the trustworthiness of LLMs and their practical utility \cite{sun2024trustllm}. However, RLHF has inherent limitations in ensuring model correctness, including insufficient evaluation of source credibility, disregard for statistical evidence, and a lack of complementary techniques to uphold the overall integrity of the content.

Future research on hallucinations in LLMs is expected to focus on optimizing model architectures and refining training methods to fundamentally reduce the occurrence of hallucinations. This may involve incorporating causal reasoning mechanisms or enhancing attention mechanisms within the models. Additionally, advancements in hallucination detection techniques are anticipated, potentially through the use of external knowledge bases and the application of contextual information to verify the accuracy of model outputs.

\subsection{Interpretability of CMR Process}

A crucial facet of advancing CMR with LLMs involves enhancing the interpretability of these models. Despite the impressive performance exhibited by LLMs in various tasks, their decision-making mechanisms frequently lack transparency, impeding their applicability in domains where interpretability is essential \cite{singh2024rethinking, creswell2022selection}, such as legal or medical contexts. 
Consequently, addressing this challenge necessitates focused research aimed at developing approaches to elucidate and visualize the reasoning processes of LLMs across different modalities. 

Recent research \cite{xue2024few,lu2022learn} has introduced effective methodologies to improve interpretability in CMR. These studies propose strategies for addressing challenges inherent in CMR, thereby enhancing the overall quality of reasoning. However, there remain opportunities for further refinement. First, it is crucial to transition from the rigid explanatory frameworks currently generated by models, advocating instead for greater flexibility to enhance the versatility of these models. Additionally, there is a pressing need to improve the precision and quality of explanations, particularly when dealing with complex issues. Lastly, there is a notable lack of consistency across different types of CMR tasks, which warrants further attention. By elevating these factors, the present interpretability of the CMR field can be significantly enhanced.

By instilling interpretability, users can gain valuable insights into the rationale behind the LLM's conclusions, fostering trust, understanding, and accountability. Moving forward, it is imperative to establish explicit interpretability standards and methodologies to facilitate the responsible and transparent deployment of LLMs in diverse cross-modal reasoning applications.

\subsection{Computational Costs and Environmental Friendliness}

LLMs not only demonstrate remarkable linguistic capabilities but also entail significant computational resource requirements. A critical challenge in this domain is the integration of diverse strategies to enhance the energy efficiency of CMR for environmental-friendliness. The primary objective, therefore, is to develop a comprehensive optimization framework for deploying LLMs in CMR tasks. This framework must effectively balance multiple considerations, including computational efficiency, parameter management, and memory utilization \cite{bai2024beyond}, ensuring that advancements in inferential capabilities do not result in disproportionate increases in resource demands compared to existing methodologies.

While prior research has focused on optimizing LLMs in general contexts \cite{ford2021portauthority, xie2019integration, wang2024model}, there remains a paucity of holistic approaches specifically tailored to enhance the performance of these models in cross-modal reasoning applications. 

Future research endeavors should prioritize the development of a comprehensive optimization framework for LLMs in CMR.  This framework should aim to balance computational efficiency, parameter management, and memory usage while maintaining or enhancing performance, with a focus on environmental friendliness. Key areas of focus should include energy efficiency enhancements, parameter reduction techniques, and task-specific optimizations. The overarching goal is to augment cross-modal capabilities without incurring substantial increases in resource demands, thereby promoting both the performance and sustainability of AI systems.

\section{Conclusion}

Within the rapidly evolving AI landscape, the integration of Large Language Models (LLMs) with cross-modal reasoning (CMR) has attracted considerable interest, highlighting the sophisticated capabilities of LLMs in language comprehension, generation, and rational processing. This survey is meticulously designed to encapsulate a detailed examination of the burgeoning research in the realm of CMR facilitated by LLMs. To elucidate the contribution of LLMs to CMR domains, we inaugurate a harmonized three-tiered taxonomy that categorizes the existing methodologies of CMR employing LLMs. This classification serves as a primer for exploring the paradigmatic approaches inherent to each delineated category. Despite the current advancements, the domain of CMR with LLMs remains immature with avenues for innovation; hence, we delineate the prevailing challenges to inspire future exploration. Through reflective discussion of these impending research vectors, we aspire to equip investigators with a profound comprehension of CMR applications within LLMs, thereby spurring continued scholarly inquiry and technological breakthroughs in this dynamic field.


\bibliographystyle{IEEEtran}
\bibliography{IEEEtran.bib}

\begin{thebibliography}{100}
\providecommand{\url}[1]{#1}
\csname url@samestyle\endcsname
\providecommand{\newblock}{\relax}
\providecommand{\bibinfo}[2]{#2}
\providecommand{\BIBentrySTDinterwordspacing}{\spaceskip=0pt\relax}
\providecommand{\BIBentryALTinterwordstretchfactor}{4}
\providecommand{\BIBentryALTinterwordspacing}{\spaceskip=\fontdimen2\font plus
\BIBentryALTinterwordstretchfactor\fontdimen3\font minus \fontdimen4\font\relax}
\providecommand{\BIBforeignlanguage}[2]{{%
\expandafter\ifx\csname l@#1\endcsname\relax
\typeout{** WARNING: IEEEtran.bst: No hyphenation pattern has been}%
\typeout{** loaded for the language `#1'. Using the pattern for}%
\typeout{** the default language instead.}%
\else
\language=\csname l@#1\endcsname
\fi
#2}}
\providecommand{\BIBdecl}{\relax}
\BIBdecl

\bibitem{openai2023gpt4}
OpenAI, ``Gpt-4 technical report,'' 2023.

\bibitem{team2023gemini}
G.~Team, R.~Anil, S.~Borgeaud, Y.~Wu, J.-B. Alayrac, J.~Yu, R.~Soricut, J.~Schalkwyk, A.~M. Dai, A.~Hauth \emph{et~al.}, ``Gemini: a family of highly capable multimodal models,'' \emph{arXiv preprint arXiv:2312.11805}, 2023.

\bibitem{anthropic2024claude}
A.~Anthropic, ``The claude 3 model family: Opus, sonnet, haiku,'' \emph{Claude-3 Model Card}, vol.~1, 2024.

\bibitem{touvron2023llama}
H.~Touvron, L.~Martin, K.~Stone, P.~Albert, A.~Almahairi, Y.~Babaei, N.~Bashlykov, S.~Batra, P.~Bhargava, S.~Bhosale \emph{et~al.}, ``Llama 2: Open foundation and fine-tuned chat models,'' \emph{arXiv preprint arXiv:2307.09288}, 2023.

\bibitem{valmeekam2022large}
K.~Valmeekam, A.~Olmo, S.~Sreedharan, and S.~Kambhampati, ``Large language models still can't plan (a benchmark for llms on planning and reasoning about change),'' in \emph{NeurIPS 2022 Foundation Models for Decision Making Workshop}, 2022.

\bibitem{wang2023can}
B.~Wang, X.~Yue, and H.~Sun, ``Can chatgpt defend its belief in truth? evaluating llm reasoning via debate,'' in \emph{Findings of the Association for Computational Linguistics: EMNLP 2023}, 2023, pp. 11\,865--11\,881.

\bibitem{laban2023summedits}
P.~Laban, W.~Kry{\'s}ci{\'n}ski, D.~Agarwal, A.~R. Fabbri, C.~Xiong, S.~Joty, and C.-S. Wu, ``Summedits: Measuring llm ability at factual reasoning through the lens of summarization,'' in \emph{Proceedings of the 2023 Conference on Empirical Methods in Natural Language Processing}, 2023, pp. 9662--9676.

\bibitem{ye2023mplug}
Q.~Ye, H.~Xu, G.~Xu, J.~Ye, M.~Yan, Y.~Zhou, J.~Wang, A.~Hu, P.~Shi, Y.~Shi \emph{et~al.}, ``mplug-owl: Modularization empowers large language models with multimodality,'' \emph{arXiv preprint arXiv:2304.14178}, 2023.

\bibitem{zheng2023minigpt}
K.~Zheng, X.~He, and X.~E. Wang, ``Minigpt-5: Interleaved vision-and-language generation via generative vokens,'' \emph{arXiv preprint arXiv:2310.02239}, 2023.

\bibitem{liu2024visual}
H.~Liu, C.~Li, Q.~Wu, and Y.~J. Lee, ``Visual instruction tuning,'' \emph{Advances in neural information processing systems}, vol.~36, 2024.

\bibitem{sloman1996empirical}
S.~A. Sloman, ``The empirical case for two systems of reasoning,'' \emph{Psychological bulletin}, vol. 119, no.~1, p.~3, 1996.

\bibitem{xue2023survey}
D.~Xue, S.~Qian, Z.~Zhou, and C.~Xu, ``A survey on interpretable cross-modal reasoning,'' \emph{arXiv preprint arXiv:2309.01955}, 2023.

\bibitem{malkinski2022review}
M.~Ma{\l}ki{\'n}ski and J.~Ma{\'n}dziuk, ``A review of emerging research directions in abstract visual reasoning,'' \emph{Information Fusion}, 2022.

\bibitem{Agrawal2015VQAVQ}
S.~Antol, A.~Agrawal, J.~Lu, M.~Mitchell, D.~Batra, C.~L. Zitnick, and D.~Parikh, ``Vqa: Visual question answering,'' in \emph{Proceedings of the IEEE international conference on computer vision}, 2015, pp. 2425--2433.

\bibitem{Hudson2019GQAAN}
D.~A. Hudson and C.~D. Manning, ``Gqa: A new dataset for real-world visual reasoning and compositional question answering,'' \emph{2019 IEEE/CVF Conference on Computer Vision and Pattern Recognition (CVPR)}, pp. 6693--6702, 2019.

\bibitem{anderson2018vision}
P.~Anderson, Q.~Wu, D.~Teney, J.~Bruce, M.~Johnson, N.~S{\"u}nderhauf, I.~Reid, S.~Gould, and A.~Van Den~Hengel, ``Vision-and-language navigation: Interpreting visually-grounded navigation instructions in real environments,'' in \emph{Proceedings of the IEEE conference on computer vision and pattern recognition}, 2018, pp. 3674--3683.

\bibitem{hao2020towards}
W.~Hao, C.~Li, X.~Li, L.~Carin, and J.~Gao, ``Towards learning a generic agent for vision-and-language navigation via pre-training,'' in \emph{Proceedings of the IEEE/CVF Conference on Computer Vision and Pattern Recognition}, 2020, pp. 13\,137--13\,146.

\bibitem{Vinyals2014ShowAT}
O.~Vinyals, A.~Toshev, S.~Bengio, and D.~Erhan, ``Show and tell: A neural image caption generator,'' \emph{2015 IEEE Conference on Computer Vision and Pattern Recognition (CVPR)}, pp. 3156--3164, 2014.

\bibitem{Xu2015ShowAA}
K.~Xu, J.~Ba, R.~Kiros, K.~Cho, A.~C. Courville, R.~Salakhutdinov, R.~S. Zemel, and Y.~Bengio, ``Show, attend and tell: Neural image caption generation with visual attention,'' in \emph{International Conference on Machine Learning}, 2015.

\bibitem{Luo2020UniViLMAU}
H.~Luo, L.~Ji, B.~Shi, H.~Huang, N.~Duan, T.~Li, X.~Chen, and M.~Zhou, ``Univilm: A unified video and language pre-training model for multimodal understanding and generation,'' \emph{ArXiv}, vol. abs/2002.06353, 2020.

\bibitem{Huang2020MultimodalPF}
G.~Huang, B.~Pang, Z.~Zhu, C.~Rivera, and R.~Soricut, ``Multimodal pretraining for dense video captioning,'' in \emph{AACL}, 2020.

\bibitem{zhu2023minigpt}
D.~Zhu, J.~Chen, X.~Shen, X.~Li, and M.~Elhoseiny, ``Minigpt-4: Enhancing vision-language understanding with advanced large language models,'' \emph{arXiv preprint arXiv:2304.10592}, 2023.

\bibitem{zhao2023chatbridge}
Z.~Zhao, L.~Guo, T.~Yue, S.~Chen, S.~Shao, X.~Zhu, Z.~Yuan, and J.~Liu, ``Chatbridge: Bridging modalities with large language model as a language catalyst,'' \emph{arXiv preprint arXiv:2305.16103}, 2023.

\bibitem{zeng2022socratic}
A.~Zeng, M.~Attarian, B.~Ichter, K.~Choromanski, A.~Wong, S.~Welker, F.~Tombari, A.~Purohit, M.~Ryoo, V.~Sindhwani \emph{et~al.}, ``Socratic models: Composing zero-shot multimodal reasoning with language,'' \emph{arXiv preprint arXiv:2204.00598}, 2022.

\bibitem{zhu2023pointclip}
X.~Zhu, R.~Zhang, B.~He, Z.~Guo, Z.~Zeng, Z.~Qin, S.~Zhang, and P.~Gao, ``Pointclip v2: Prompting clip and gpt for powerful 3d open-world learning,'' in \emph{Proceedings of the IEEE/CVF International Conference on Computer Vision}, 2023, pp. 2639--2650.

\bibitem{You2023IdealGPTID}
H.~You, R.~Sun, Z.~Wang, L.~Chen, G.~Wang, H.~A. Ayyubi, K.-W. Chang, and S.-F. Chang, ``Idealgpt: Iteratively decomposing vision and language reasoning via large language models,'' \emph{ArXiv}, vol. abs/2305.14985, 2023.

\bibitem{guo2023viewrefer}
Z.~Guo, Y.~Tang, R.~Zhang, D.~Wang, Z.~Wang, B.~Zhao, and X.~Li, ``Viewrefer: Grasp the multi-view knowledge for 3d visual grounding,'' in \emph{Proceedings of the IEEE/CVF International Conference on Computer Vision}, 2023, pp. 15\,372--15\,383.

\bibitem{zhu2022pointclip}
X.~Zhu, R.~Zhang, B.~He, Z.~Zeng, S.~Zhang, and P.~Gao, ``Pointclip v2: Adapting clip for powerful 3d open-world learning,'' \emph{arXiv preprint arXiv:2211.11682}, 2022.

\bibitem{yang2023mm}
Z.~Yang, L.~Li, J.~Wang, K.~Lin, E.~Azarnasab, F.~Ahmed, Z.~Liu, C.~Liu, M.~Zeng, and L.~Wang, ``Mm-react: Prompting chatgpt for multimodal reasoning and action,'' \emph{arXiv preprint arXiv:2303.11381}, 2023.

\bibitem{huang2022inner}
W.~Huang, F.~Xia, T.~Xiao, H.~Chan, J.~Liang, P.~Florence, A.~Zeng, J.~Tompson, I.~Mordatch, Y.~Chebotar \emph{et~al.}, ``Inner monologue: Embodied reasoning through planning with language models,'' \emph{arXiv preprint arXiv:2207.05608}, 2022.

\bibitem{singh2023progprompt}
I.~Singh, V.~Blukis, A.~Mousavian, A.~Goyal, D.~Xu, J.~Tremblay, D.~Fox, J.~Thomason, and A.~Garg, ``Progprompt: Generating situated robot task plans using large language models,'' in \emph{2023 IEEE International Conference on Robotics and Automation (ICRA)}.\hskip 1em plus 0.5em minus 0.4em\relax IEEE, 2023, pp. 11\,523--11\,530.

\bibitem{gupta2023visual}
T.~Gupta and A.~Kembhavi, ``Visual programming: Compositional visual reasoning without training,'' in \emph{Proceedings of the IEEE/CVF Conference on Computer Vision and Pattern Recognition}, 2023, pp. 14\,953--14\,962.

\bibitem{zhang2023prompt}
R.~Zhang, X.~Hu, B.~Li, S.~Huang, H.~Deng, Y.~Qiao, P.~Gao, and H.~Li, ``Prompt, generate, then cache: Cascade of foundation models makes strong few-shot learners,'' in \emph{Proceedings of the IEEE/CVF Conference on Computer Vision and Pattern Recognition}, 2023, pp. 15\,211--15\,222.

\bibitem{pratt2023does}
S.~Pratt, I.~Covert, R.~Liu, and A.~Farhadi, ``What does a platypus look like? generating customized prompts for zero-shot image classification,'' in \emph{Proceedings of the IEEE/CVF International Conference on Computer Vision}, 2023, pp. 15\,691--15\,701.

\bibitem{shen2023hugginggpt}
Y.~Shen, K.~Song, X.~Tan, D.~Li, W.~Lu, and Y.~Zhuang, ``Hugginggpt: Solving ai tasks with chatgpt and its friends in huggingface,'' \emph{arXiv preprint arXiv:2303.17580}, 2023.

\bibitem{berger2017spatio}
C.~Berger, J.~Rosentreter, M.~Voltersen, C.~Baumgart, C.~Schmullius, and S.~Hese, ``Spatio-temporal analysis of the relationship between 2d/3d urban site characteristics and land surface temperature,'' \emph{Remote sensing of environment}, vol. 193, pp. 225--243, 2017.

\bibitem{zhao2023survey}
W.~X. Zhao, K.~Zhou, J.~Li, T.~Tang, X.~Wang, Y.~Hou, Y.~Min, B.~Zhang, J.~Zhang, Z.~Dong \emph{et~al.}, ``A survey of large language models,'' \emph{arXiv preprint arXiv:2303.18223}, 2023.

\bibitem{huang2022towards}
J.~Huang and K.~C.-C. Chang, ``Towards reasoning in large language models: A survey,'' \emph{arXiv preprint arXiv:2212.10403}, 2022.

\bibitem{wang2023aligning}
Y.~Wang, W.~Zhong, L.~Li, F.~Mi, X.~Zeng, W.~Huang, L.~Shang, X.~Jiang, and Q.~Liu, ``Aligning large language models with human: A survey,'' \emph{arXiv preprint arXiv:2307.12966}, 2023.

\bibitem{qiao2022reasoning}
S.~Qiao, Y.~Ou, N.~Zhang, X.~Chen, Y.~Yao, S.~Deng, C.~Tan, F.~Huang, and H.~Chen, ``Reasoning with language model prompting: A survey,'' \emph{arXiv preprint arXiv:2212.09597}, 2022.

\bibitem{yin2023survey}
S.~Yin, C.~Fu, S.~Zhao, K.~Li, X.~Sun, T.~Xu, and E.~Chen, ``A survey on multimodal large language models,'' \emph{arXiv preprint arXiv:2306.13549}, 2023.

\bibitem{jain2024vcoder}
J.~Jain, J.~Yang, and H.~Shi, ``Vcoder: Versatile vision encoders for multimodal large language models,'' in \emph{Proceedings of the IEEE/CVF Conference on Computer Vision and Pattern Recognition}, 2024, pp. 27\,992--28\,002.

\bibitem{zhu2024beyond}
L.~Zhu, F.~Wei, and Y.~Lu, ``Beyond text: Frozen large language models in visual signal comprehension,'' in \emph{Proceedings of the IEEE/CVF Conference on Computer Vision and Pattern Recognition}, 2024, pp. 27\,047--27\,057.

\bibitem{liu2024bt}
R.~Liu, C.~Li, Y.~Ge, T.~H. Li, Y.~Shan, and G.~Li, ``Bt-adapter: Video conversation is feasible without video instruction tuning,'' in \emph{Proceedings of the IEEE/CVF Conference on Computer Vision and Pattern Recognition}, 2024, pp. 13\,658--13\,667.

\bibitem{zhang2023multi}
Y.~Zhang, L.~Cui, D.~Cai, X.~Huang, T.~Fang, and W.~Bi, ``Multi-task instruction tuning of llama for specific scenarios: A preliminary study on writing assistance,'' \emph{arXiv preprint arXiv:2305.13225}, 2023.

\bibitem{chen2024lion}
G.~Chen, L.~Shen, R.~Shao, X.~Deng, and L.~Nie, ``Lion: Empowering multimodal large language model with dual-level visual knowledge,'' in \emph{Proceedings of the IEEE/CVF Conference on Computer Vision and Pattern Recognition}, 2024, pp. 26\,540--26\,550.

\bibitem{pramanick2024jack}
S.~Pramanick, G.~Han, R.~Hou, S.~Nag, S.-N. Lim, N.~Ballas, Q.~Wang, R.~Chellappa, and A.~Almahairi, ``Jack of all tasks master of many: Designing general-purpose coarse-to-fine vision-language model,'' in \emph{Proceedings of the IEEE/CVF Conference on Computer Vision and Pattern Recognition}, 2024, pp. 14\,076--14\,088.

\bibitem{chen2024far}
Z.~Chen, W.~Wang, H.~Tian, S.~Ye, Z.~Gao, E.~Cui, W.~Tong, K.~Hu, J.~Luo, Z.~Ma \emph{et~al.}, ``How far are we to gpt-4v? closing the gap to commercial multimodal models with open-source suites,'' \emph{arXiv preprint arXiv:2404.16821}, 2024.

\bibitem{jin2024chat}
P.~Jin, R.~Takanobu, W.~Zhang, X.~Cao, and L.~Yuan, ``Chat-univi: Unified visual representation empowers large language models with image and video understanding,'' in \emph{Proceedings of the IEEE/CVF Conference on Computer Vision and Pattern Recognition}, 2024, pp. 13\,700--13\,710.

\bibitem{zhang2023gpt4roi}
S.~Zhang, P.~Sun, S.~Chen, M.~Xiao, W.~Shao, W.~Zhang, Y.~Liu, K.~Chen, and P.~Luo, ``Gpt4roi: Instruction tuning large language model on region-of-interest,'' \emph{arXiv preprint arXiv:2307.03601}, 2023.

\bibitem{dong2024internlm}
X.~Dong, P.~Zhang, Y.~Zang, Y.~Cao, B.~Wang, L.~Ouyang, X.~Wei, S.~Zhang, H.~Duan, M.~Cao \emph{et~al.}, ``Internlm-xcomposer2: Mastering free-form text-image composition and comprehension in vision-language large model,'' \emph{arXiv preprint arXiv:2401.16420}, 2024.

\bibitem{yao2024minicpm}
Y.~Yao, T.~Yu, A.~Zhang, C.~Wang, J.~Cui, H.~Zhu, T.~Cai, H.~Li, W.~Zhao, Z.~He \emph{et~al.}, ``Minicpm-v: A gpt-4v level mllm on your phone,'' \emph{arXiv preprint arXiv:2408.01800}, 2024.

\bibitem{xue2024xgen}
L.~Xue, M.~Shu, A.~Awadalla, J.~Wang, A.~Yan, S.~Purushwalkam, H.~Zhou, V.~Prabhu, Y.~Dai, M.~S. Ryoo \emph{et~al.}, ``xgen-mm (blip-3): A family of open large multimodal models,'' \emph{arXiv preprint arXiv:2408.08872}, 2024.

\bibitem{radford2018improving}
A.~Radford, K.~Narasimhan, T.~Salimans, I.~Sutskever \emph{et~al.}, ``Improving language understanding by generative pre-training,'' 2018.

\bibitem{NEURIPS20201457c0d6}
T.~Brown, B.~Mann, N.~Ryder, M.~Subbiah, J.~D. Kaplan, P.~Dhariwal, A.~Neelakantan, P.~Shyam, G.~Sastry, A.~Askell \emph{et~al.}, ``Language models are few-shot learners,'' \emph{Advances in neural information processing systems}, vol.~33, pp. 1877--1901, 2020.

\bibitem{radford2019language}
A.~Radford, J.~Wu, R.~Child, D.~Luan, D.~Amodei, I.~Sutskever \emph{et~al.}, ``Language models are unsupervised multitask learners,'' \emph{OpenAI blog}, vol.~1, no.~8, p.~9, 2019.

\bibitem{devlin2018bert}
J.~Devlin, M.-W. Chang, K.~Lee, and K.~Toutanova, ``Bert: Pre-training of deep bidirectional transformers for language understanding,'' \emph{arXiv preprint arXiv:1810.04805}, 2018.

\bibitem{liu2019roberta}
Y.~Liu, M.~Ott, N.~Goyal, J.~Du, M.~Joshi, D.~Chen, O.~Levy, M.~Lewis, L.~Zettlemoyer, and V.~Stoyanov, ``Roberta: A robustly optimized bert pretraining approach,'' \emph{arXiv preprint arXiv:1907.11692}, 2019.

\bibitem{lewis2019bart}
M.~Lewis, Y.~Liu, N.~Goyal, M.~Ghazvininejad, A.~Mohamed, O.~Levy, V.~Stoyanov, and L.~Zettlemoyer, ``Bart: Denoising sequence-to-sequence pre-training for natural language generation, translation, and comprehension,'' \emph{arXiv preprint arXiv:1910.13461}, 2019.

\bibitem{Wang2023CaptionAI}
T.~Wang, J.~Zhang, J.~Fei, Y.~Ge, H.~Zheng, Y.~Tang, Z.~Li, M.~Gao, S.~Zhao, Y.~Shan, and F.~Zheng, ``Caption anything: Interactive image description with diverse multimodal controls,'' \emph{ArXiv}, vol. abs/2305.02677, 2023.

\bibitem{lu2023chameleon}
P.~Lu, B.~Peng, H.~Cheng, M.~Galley, K.-W. Chang, Y.~N. Wu, S.-C. Zhu, and J.~Gao, ``Chameleon: Plug-and-play compositional reasoning with large language models,'' \emph{arXiv preprint arXiv:2304.09842}, 2023.

\bibitem{yang2023gpt4tools}
R.~Yang, L.~Song, Y.~Li, S.~Zhao, Y.~Ge, X.~Li, and Y.~Shan, ``Gpt4tools: Teaching large language model to use tools via self-instruction,'' \emph{arXiv preprint arXiv:2305.18752}, 2023.

\bibitem{Zhu2023ChatGPTAB}
D.~Zhu, J.~Chen, K.~Haydarov, X.~Shen, W.~Zhang, and M.~Elhoseiny, ``Chatgpt asks, blip-2 answers: Automatic questioning towards enriched visual descriptions,'' \emph{ArXiv}, vol. abs/2303.06594, 2023.

\bibitem{chen2023shikra}
K.~Chen, Z.~Zhang, W.~Zeng, R.~Zhang, F.~Zhu, and R.~Zhao, ``Shikra: Unleashing multimodal llm's referential dialogue magic,'' \emph{arXiv preprint arXiv:2306.15195}, 2023.

\bibitem{wang2023wall}
T.~Wang, Y.~Li, H.~Lin, X.~Xue, and Y.~Fu, ``Wall-e: Embodied robotic waiter load lifting with large language model,'' \emph{arXiv preprint arXiv:2308.15962}, 2023.

\bibitem{10.1145/3560815}
P.~Liu, W.~Yuan, J.~Fu, Z.~Jiang, H.~Hayashi, and G.~Neubig, ``Pre-train, prompt, and predict: A systematic survey of prompting methods in natural language processing,'' \emph{ACM Comput. Surv.}, vol.~55, no.~9, jan 2023.

\bibitem{Lester2021ThePO}
B.~Lester, R.~Al-Rfou, and N.~Constant, ``The power of scale for parameter-efficient prompt tuning,'' in \emph{Conference on Empirical Methods in Natural Language Processing}, 2021.

\bibitem{reynolds2021prompt}
L.~Reynolds and K.~McDonell, ``Prompt programming for large language models: Beyond the few-shot paradigm,'' 2021.

\bibitem{Argyle2022AnIA}
L.~P. Argyle, E.~Busby, N.~Fulda, J.~R. Gubler, C.~Rytting, T.~Sorensen, and D.~Wingate, ``An information-theoretic approach to prompt engineering without ground truth labels,'' \emph{Political Analysis}, vol.~31, pp. 337 -- 351, 2022.

\bibitem{gui2021kat}
L.~Gui, B.~Wang, Q.~Huang, A.~Hauptmann, Y.~Bisk, and J.~Gao, ``Kat: A knowledge augmented transformer for vision-and-language,'' \emph{arXiv preprint arXiv:2112.08614}, 2021.

\bibitem{lin2022revive}
Y.~Lin, Y.~Xie, D.~Chen, Y.~Xu, C.~Zhu, and L.~Yuan, ``Revive: Regional visual representation matters in knowledge-based visual question answering,'' \emph{Advances in Neural Information Processing Systems}, vol.~35, pp. 10\,560--10\,571, 2022.

\bibitem{wu2023visual}
C.~Wu, S.~Yin, W.~Qi, X.~Wang, Z.~Tang, and N.~Duan, ``Visual chatgpt: Talking, drawing and editing with visual foundation models,'' \emph{arXiv preprint arXiv:2303.04671}, 2023.

\bibitem{driess2023palm}
D.~Driess, F.~Xia, M.~S. Sajjadi, C.~Lynch, A.~Chowdhery, B.~Ichter, A.~Wahid, J.~Tompson, Q.~Vuong, T.~Yu \emph{et~al.}, ``Palm-e: An embodied multimodal language model,'' \emph{arXiv preprint arXiv:2303.03378}, 2023.

\bibitem{Chowdhery2022PaLMSL}
A.~Chowdhery, S.~Narang, J.~Devlin, M.~Bosma, G.~Mishra, A.~Roberts, P.~Barham, H.~W. Chung, C.~Sutton, S.~Gehrmann \emph{et~al.}, ``Palm: Scaling language modeling with pathways,'' \emph{arXiv preprint arXiv:2204.02311}, 2022.

\bibitem{cho2021unifying}
J.~Cho, J.~Lei, H.~Tan, and M.~Bansal, ``Unifying vision-and-language tasks via text generation,'' in \emph{International Conference on Machine Learning}.\hskip 1em plus 0.5em minus 0.4em\relax PMLR, 2021, pp. 1931--1942.

\bibitem{raffel2020exploring}
C.~Raffel, N.~Shazeer, A.~Roberts, K.~Lee, S.~Narang, M.~Matena, Y.~Zhou, W.~Li, and P.~J. Liu, ``Exploring the limits of transfer learning with a unified text-to-text transformer,'' \emph{The Journal of Machine Learning Research}, vol.~21, no.~1, pp. 5485--5551, 2020.

\bibitem{ren2015faster}
S.~Ren, K.~He, R.~Girshick, and J.~Sun, ``Faster r-cnn: Towards real-time object detection with region proposal networks,'' \emph{Advances in neural information processing systems}, vol.~28, 2015.

\bibitem{li2021prefix}
X.~L. Li and P.~Liang, ``Prefix-tuning: Optimizing continuous prompts for generation,'' \emph{arXiv preprint arXiv:2101.00190}, 2021.

\bibitem{li2023blip}
J.~Li, D.~Li, S.~Savarese, and S.~Hoi, ``Blip-2: Bootstrapping language-image pre-training with frozen image encoders and large language models,'' \emph{arXiv preprint arXiv:2301.12597}, 2023.

\bibitem{zhang2022opt}
S.~Zhang, S.~Roller, N.~Goyal, M.~Artetxe, M.~Chen, S.~Chen, C.~Dewan, M.~Diab, X.~Li, X.~V. Lin \emph{et~al.}, ``Opt: Open pre-trained transformer language models,'' \emph{arXiv preprint arXiv:2205.01068}, 2022.

\bibitem{Chung2022ScalingIL}
H.~W. Chung, L.~Hou, S.~Longpre, B.~Zoph, Y.~Tay, W.~Fedus, Y.~Li, X.~Wang, M.~Dehghani, S.~Brahma \emph{et~al.}, ``Scaling instruction-finetuned language models,'' \emph{arXiv preprint arXiv:2210.11416}, 2022.

\bibitem{zhang2023transfer}
A.~Zhang, H.~Fei, Y.~Yao, W.~Ji, L.~Li, Z.~Liu, and T.-S. Chua, ``Transfer visual prompt generator across llms,'' \emph{arXiv preprint arXiv:2305.01278}, 2023.

\bibitem{shukor2023ep}
M.~Shukor, C.~Dancette, and M.~Cord, ``ep-alm: Efficient perceptual augmentation of language models,'' \emph{arXiv preprint arXiv:2303.11403}, 2023.

\bibitem{dosovitskiy2020image}
A.~Dosovitskiy, L.~Beyer, A.~Kolesnikov, D.~Weissenborn, X.~Zhai, T.~Unterthiner, M.~Dehghani, M.~Minderer, G.~Heigold, S.~Gelly \emph{et~al.}, ``An image is worth 16x16 words: Transformers for image recognition at scale,'' \emph{arXiv preprint arXiv:2010.11929}, 2020.

\bibitem{Rose2023VisualCO}
D.~P. Rose, V.~Himakunthala, A.~Ouyang, R.~He, A.~Mei, Y.~Lu, M.~S. Saxon, C.~Sonar, D.~Mirza, and W.~Y. Wang, ``Visual chain of thought: Bridging logical gaps with multimodal infillings,'' \emph{ArXiv}, vol. abs/2305.02317, 2023.

\bibitem{xu2022multiinstruct}
Z.~Xu, Y.~Shen, and L.~Huang, ``Multiinstruct: Improving multi-modal zero-shot learning via instruction tuning,'' \emph{arXiv preprint arXiv:2212.10773}, 2022.

\bibitem{wang2022ofa}
P.~Wang, A.~Yang, R.~Men, J.~Lin, S.~Bai, Z.~Li, J.~Ma, C.~Zhou, J.~Zhou, and H.~Yang, ``Ofa: Unifying architectures, tasks, and modalities through a simple sequence-to-sequence learning framework,'' in \emph{International Conference on Machine Learning}.\hskip 1em plus 0.5em minus 0.4em\relax PMLR, 2022, pp. 23\,318--23\,340.

\bibitem{mishra2021cross}
S.~Mishra, D.~Khashabi, C.~Baral, and H.~Hajishirzi, ``Cross-task generalization via natural language crowdsourcing instructions,'' \emph{arXiv preprint arXiv:2104.08773}, 2021.

\bibitem{dai2023instructblip}
W.~Dai, J.~Li, D.~Li, A.~M.~H. Tiong, J.~Zhao, W.~Wang, B.~Li, P.~Fung, and S.~Hoi, ``Instructblip: Towards general-purpose vision-language models with instruction tuning,'' 2023.

\bibitem{liu2024improved}
H.~Liu, C.~Li, Y.~Li, and Y.~J. Lee, ``Improved baselines with visual instruction tuning,'' in \emph{Proceedings of the IEEE/CVF Conference on Computer Vision and Pattern Recognition}, 2024, pp. 26\,296--26\,306.

\bibitem{maaz2023video}
M.~Maaz, H.~Rasheed, S.~Khan, and F.~S. Khan, ``Video-chatgpt: Towards detailed video understanding via large vision and language models,'' \emph{arXiv preprint arXiv:2306.05424}, 2023.

\bibitem{li2023llava}
C.~Li, C.~Wong, S.~Zhang, N.~Usuyama, H.~Liu, J.~Yang, T.~Naumann, H.~Poon, and J.~Gao, ``Llava-med: Training a large language-and-vision assistant for biomedicine in one day,'' \emph{arXiv preprint arXiv:2306.00890}, 2023.

\bibitem{zhang2023pmc}
X.~Zhang, C.~Wu, Z.~Zhao, W.~Lin, Y.~Zhang, Y.~Wang, and W.~Xie, ``Pmc-vqa: Visual instruction tuning for medical visual question answering,'' \emph{arXiv preprint arXiv:2305.10415}, 2023.

\bibitem{hu2021lora}
E.~J. Hu, Y.~Shen, P.~Wallis, Z.~Allen-Zhu, Y.~Li, S.~Wang, L.~Wang, and W.~Chen, ``Lora: Low-rank adaptation of large language models,'' \emph{arXiv preprint arXiv:2106.09685}, 2021.

\bibitem{chiang2023vicuna}
W.-L. Chiang, Z.~Li, Z.~Lin, Y.~Sheng, Z.~Wu, H.~Zhang, L.~Zheng, S.~Zhuang, Y.~Zhuang, J.~E. Gonzalez \emph{et~al.}, ``Vicuna: An open-source chatbot impressing gpt-4 with 90\%* chatgpt quality,'' \emph{See https://vicuna. lmsys. org (accessed 14 April 2023)}, 2023.

\bibitem{Li2023VideoChatCV}
K.~Li, Y.~He, Y.~Wang, Y.~Li, W.~Wang, P.~Luo, Y.~Wang, L.~Wang, and Y.~Qiao, ``Videochat: Chat-centric video understanding,'' \emph{ArXiv}, vol. abs/2305.06355, 2023.

\bibitem{luo2023cheap}
G.~Luo, Y.~Zhou, T.~Ren, S.~Chen, X.~Sun, and R.~Ji, ``Cheap and quick: Efficient vision-language instruction tuning for large language models,'' \emph{arXiv preprint arXiv:2305.15023}, 2023.

\bibitem{zhang2023video}
H.~Zhang, X.~Li, and L.~Bing, ``Video-llama: An instruction-tuned audio-visual language model for video understanding,'' \emph{arXiv preprint arXiv:2306.02858}, 2023.

\bibitem{pi2023detgpt}
R.~Pi, J.~Gao, S.~Diao, R.~Pan, H.~Dong, J.~Zhang, L.~Yao, J.~Han, H.~Xu, and L.~K.~T. Zhang, ``Detgpt: Detect what you need via reasoning,'' \emph{arXiv preprint arXiv:2305.14167}, 2023.

\bibitem{lyu2023macaw}
C.~Lyu, M.~Wu, L.~Wang, X.~Huang, B.~Liu, Z.~Du, S.~Shi, and Z.~Tu, ``Macaw-llm: Multi-modal language modeling with image, audio, video, and text integration,'' \emph{arXiv preprint arXiv:2306.09093}, 2023.

\bibitem{zhang2023llama}
R.~Zhang, J.~Han, A.~Zhou, X.~Hu, S.~Yan, P.~Lu, H.~Li, P.~Gao, and Y.~Qiao, ``Llama-adapter: Efficient fine-tuning of language models with zero-init attention,'' \emph{arXiv preprint arXiv:2303.16199}, 2023.

\bibitem{liu2022few}
H.~Liu, D.~Tam, M.~Muqeeth, J.~Mohta, T.~Huang, M.~Bansal, and C.~A. Raffel, ``Few-shot parameter-efficient fine-tuning is better and cheaper than in-context learning,'' \emph{Advances in Neural Information Processing Systems}, vol.~35, pp. 1950--1965, 2022.

\bibitem{gao2023llama}
P.~Gao, J.~Han, R.~Zhang, Z.~Lin, S.~Geng, A.~Zhou, W.~Zhang, P.~Lu, C.~He, X.~Yue \emph{et~al.}, ``Llama-adapter v2: Parameter-efficient visual instruction model,'' \emph{arXiv preprint arXiv:2304.15010}, 2023.

\bibitem{chen2023minigpt}
J.~Chen, D.~Zhu, X.~Shen, X.~Li, Z.~Liu, P.~Zhang, R.~Krishnamoorthi, V.~Chandra, Y.~Xiong, and M.~Elhoseiny, ``Minigpt-v2: large language model as a unified interface for vision-language multi-task learning,'' \emph{arXiv preprint arXiv:2310.09478}, 2023.

\bibitem{su2023pandagpt}
Y.~Su, T.~Lan, H.~Li, J.~Xu, Y.~Wang, and D.~Cai, ``Pandagpt: One model to instruction-follow them all,'' \emph{arXiv preprint arXiv:2305.16355}, 2023.

\bibitem{girdhar2023imagebind}
R.~Girdhar, A.~El-Nouby, Z.~Liu, M.~Singh, K.~V. Alwala, A.~Joulin, and I.~Misra, ``Imagebind: One embedding space to bind them all,'' in \emph{Proceedings of the IEEE/CVF Conference on Computer Vision and Pattern Recognition}, 2023, pp. 15\,180--15\,190.

\bibitem{gong2023multimodal}
T.~Gong, C.~Lyu, S.~Zhang, Y.~Wang, M.~Zheng, Q.~Zhao, K.~Liu, W.~Zhang, P.~Luo, and K.~Chen, ``Multimodal-gpt: A vision and language model for dialogue with humans,'' \emph{arXiv preprint arXiv:2305.04790}, 2023.

\bibitem{li2023m}
L.~Li, Y.~Yin, S.~Li, L.~Chen, P.~Wang, S.~Ren, M.~Li, Y.~Yang, J.~Xu, X.~Sun \emph{et~al.}, ``M$^{3}$it: A large-scale dataset towards multi-modal multilingual instruction tuning,'' \emph{arXiv preprint arXiv:2306.04387}, 2023.

\bibitem{zhang2022fengshenbang}
J.~Zhang, R.~Gan, J.~Wang, Y.~Zhang, L.~Zhang, P.~Yang, X.~Gao, Z.~Wu, X.~Dong, J.~He \emph{et~al.}, ``Fengshenbang 1.0: Being the foundation of chinese cognitive intelligence,'' \emph{arXiv preprint arXiv:2209.02970}, 2022.

\bibitem{chen2023visual}
D.~Chen, J.~Liu, W.~Dai, and B.~Wang, ``Visual instruction tuning with polite flamingo,'' \emph{arXiv preprint arXiv:2307.01003}, 2023.

\bibitem{wu2023next}
S.~Wu, H.~Fei, L.~Qu, W.~Ji, and T.-S. Chua, ``Next-gpt: Any-to-any multimodal llm,'' \emph{arXiv preprint arXiv:2309.05519}, 2023.

\bibitem{zhao2023chatspot}
L.~Zhao, E.~Yu, Z.~Ge, J.~Yang, H.~Wei, H.~Zhou, J.~Sun, Y.~Peng, R.~Dong, C.~Han \emph{et~al.}, ``Chatspot: Bootstrapping multimodal llms via precise referring instruction tuning,'' \emph{arXiv preprint arXiv:2307.09474}, 2023.

\bibitem{hu2023bliva}
W.~Hu, Y.~Xu, Y.~Li, W.~Li, Z.~Chen, and Z.~Tu, ``Bliva: A simple multimodal llm for better handling of text-rich visual questions,'' \emph{arXiv preprint arXiv:2308.09936}, 2023.

\bibitem{zhao2023bubogpt}
Y.~Zhao, Z.~Lin, D.~Zhou, Z.~Huang, J.~Feng, and B.~Kang, ``Bubogpt: Enabling visual grounding in multi-modal llms,'' \emph{arXiv preprint arXiv:2307.08581}, 2023.

\bibitem{wang2023visionllm}
W.~Wang, Z.~Chen, X.~Chen, J.~Wu, X.~Zhu, G.~Zeng, P.~Luo, T.~Lu, J.~Zhou, Y.~Qiao \emph{et~al.}, ``Visionllm: Large language model is also an open-ended decoder for vision-centric tasks,'' \emph{arXiv preprint arXiv:2305.11175}, 2023.

\bibitem{yin2023lamm}
Z.~Yin, J.~Wang, J.~Cao, Z.~Shi, D.~Liu, M.~Li, L.~Sheng, L.~Bai, X.~Huang, Z.~Wang \emph{et~al.}, ``Lamm: Language-assisted multi-modal instruction-tuning dataset, framework, and benchmark,'' \emph{arXiv preprint arXiv:2306.06687}, 2023.

\bibitem{bai2023qwen}
J.~Bai, S.~Bai, S.~Yang, S.~Wang, S.~Tan, P.~Wang, J.~Lin, C.~Zhou, and J.~Zhou, ``Qwen-vl: A frontier large vision-language model with versatile abilities,'' \emph{arXiv preprint arXiv:2308.12966}, 2023.

\bibitem{ye2023mplug2}
Q.~Ye, H.~Xu, J.~Ye, M.~Yan, H.~Liu, Q.~Qian, J.~Zhang, F.~Huang, and J.~Zhou, ``mplug-owl2: Revolutionizing multi-modal large language model with modality collaboration,'' \emph{arXiv preprint arXiv:2311.04257}, 2023.

\bibitem{ye2024mplug}
J.~Ye, H.~Xu, H.~Liu, A.~Hu, M.~Yan, Q.~Qian, J.~Zhang, F.~Huang, and J.~Zhou, ``mplug-owl3: Towards long image-sequence understanding in multi-modal large language models,'' \emph{arXiv preprint arXiv:2408.04840}, 2024.

\bibitem{chen2023x}
F.~Chen, M.~Han, H.~Zhao, Q.~Zhang, J.~Shi, S.~Xu, and B.~Xu, ``X-llm: Bootstrapping advanced large language models by treating multi-modalities as foreign languages,'' \emph{arXiv preprint arXiv:2305.04160}, 2023.

\bibitem{alayrac2022flamingo}
J.-B. Alayrac, J.~Donahue, P.~Luc, A.~Miech, I.~Barr, Y.~Hasson, K.~Lenc, A.~Mensch, K.~Millican, M.~Reynolds \emph{et~al.}, ``Flamingo: a visual language model for few-shot learning,'' \emph{Advances in Neural Information Processing Systems}, vol.~35, pp. 23\,716--23\,736, 2022.

\bibitem{li2023empowering}
J.~Li, K.~Pan, Z.~Ge, M.~Gao, H.~Zhang, W.~Ji, W.~Zhang, T.-S. Chua, S.~Tang, and Y.~Zhuang, ``Empowering vision-language models to follow interleaved vision-language instructions,'' \emph{arXiv preprint arXiv:2308.04152}, 2023.

\bibitem{eichenberg2021magma}
C.~Eichenberg, S.~Black, S.~Weinbach, L.~Parcalabescu, and A.~Frank, ``Magma--multimodal augmentation of generative models through adapter-based finetuning,'' \emph{arXiv preprint arXiv:2112.05253}, 2021.

\bibitem{wang2021gpt}
B.~Wang and A.~Komatsuzaki, ``Gpt-j-6b: A 6 billion parameter autoregressive language model,'' 2021.

\bibitem{liu2023prismer}
S.~Liu, L.~Fan, E.~Johns, Z.~Yu, C.~Xiao, and A.~Anandkumar, ``Prismer: A vision-language model with an ensemble of experts,'' \emph{arXiv preprint arXiv:2303.02506}, 2023.

\bibitem{chen2022pali}
X.~Chen, X.~Wang, S.~Changpinyo, A.~Piergiovanni, P.~Padlewski, D.~Salz, S.~Goodman, A.~Grycner, B.~Mustafa, L.~Beyer \emph{et~al.}, ``Pali: A jointly-scaled multilingual language-image model,'' \emph{arXiv preprint arXiv:2209.06794}, 2022.

\bibitem{zhai2022scaling}
X.~Zhai, A.~Kolesnikov, N.~Houlsby, and L.~Beyer, ``Scaling vision transformers,'' in \emph{Proceedings of the IEEE/CVF Conference on Computer Vision and Pattern Recognition}, 2022, pp. 12\,104--12\,113.

\bibitem{xue2020mt5}
L.~Xue, N.~Constant, A.~Roberts, M.~Kale, R.~Al-Rfou, A.~Siddhant, A.~Barua, and C.~Raffel, ``mt5: A massively multilingual pre-trained text-to-text transformer,'' \emph{arXiv preprint arXiv:2010.11934}, 2020.

\bibitem{radford2021learning}
A.~Radford, J.~W. Kim, C.~Hallacy, A.~Ramesh, G.~Goh, S.~Agarwal, G.~Sastry, A.~Askell, P.~Mishkin, J.~Clark \emph{et~al.}, ``Learning transferable visual models from natural language supervision,'' in \emph{International conference on machine learning}.\hskip 1em plus 0.5em minus 0.4em\relax PMLR, 2021, pp. 8748--8763.

\bibitem{bertasius2021space}
G.~Bertasius, H.~Wang, and L.~Torresani, ``Is space-time attention all you need for video understanding?'' in \emph{ICML}, vol.~2, no.~3, 2021, p.~4.

\bibitem{zhao2023learning}
Y.~Zhao, I.~Misra, P.~Kr{\"a}henb{\"u}hl, and R.~Girdhar, ``Learning video representations from large language models,'' in \emph{Proceedings of the IEEE/CVF Conference on Computer Vision and Pattern Recognition}, 2023, pp. 6586--6597.

\bibitem{yan2023videococa}
S.~Yan, T.~Zhu, Z.~Wang, Y.~Cao, M.~Zhang, S.~Ghosh, Y.~Wu, and J.~Yu, ``Videococa: Video-text modeling with zero-shot transfer from contrastive captioners,'' 2023.

\bibitem{yu2022coca}
J.~Yu, Z.~Wang, V.~Vasudevan, L.~Yeung, M.~Seyedhosseini, and Y.~Wu, ``Coca: Contrastive captioners are image-text foundation models,'' \emph{arXiv preprint arXiv:2205.01917}, 2022.

\bibitem{awadalla2023openflamingo}
A.~Awadalla, I.~Gao, J.~Gardner, J.~Hessel, Y.~Hanafy, W.~Zhu, K.~Marathe, Y.~Bitton, S.~Gadre, S.~Sagawa \emph{et~al.}, ``Openflamingo: An open-source framework for training large autoregressive vision-language models,'' \emph{arXiv preprint arXiv:2308.01390}, 2023.

\bibitem{li2023mimic}
B.~Li, Y.~Zhang, L.~Chen, J.~Wang, F.~Pu, J.~Yang, C.~Li, and Z.~Liu, ``Mimic-it: Multi-modal in-context instruction tuning,'' \emph{arXiv preprint arXiv:2306.05425}, 2023.

\bibitem{reed2022generalist}
S.~Reed, K.~Zolna, E.~Parisotto, S.~G. Colmenarejo, A.~Novikov, G.~Barth-Maron, M.~Gimenez, Y.~Sulsky, J.~Kay, J.~T. Springenberg \emph{et~al.}, ``A generalist agent,'' \emph{arXiv preprint arXiv:2205.06175}, 2022.

\bibitem{chen2023pali}
X.~Chen, J.~Djolonga, P.~Padlewski, B.~Mustafa, S.~Changpinyo, J.~Wu, C.~R. Ruiz, S.~Goodman, X.~Wang, Y.~Tay \emph{et~al.}, ``Pali-x: On scaling up a multilingual vision and language model,'' \emph{arXiv preprint arXiv:2305.18565}, 2023.

\bibitem{dehghani2023scaling}
M.~Dehghani, J.~Djolonga, B.~Mustafa, P.~Padlewski, J.~Heek, J.~Gilmer, A.~P. Steiner, M.~Caron, R.~Geirhos, I.~Alabdulmohsin \emph{et~al.}, ``Scaling vision transformers to 22 billion parameters,'' in \emph{International Conference on Machine Learning}.\hskip 1em plus 0.5em minus 0.4em\relax PMLR, 2023, pp. 7480--7512.

\bibitem{tay2022ul2}
Y.~Tay, M.~Dehghani, V.~Q. Tran, X.~Garcia, J.~Wei, X.~Wang, H.~W. Chung, D.~Bahri, T.~Schuster, S.~Zheng \emph{et~al.}, ``Ul2: Unifying language learning paradigms,'' in \emph{The Eleventh International Conference on Learning Representations}, 2022.

\bibitem{chen2023cosa}
S.~Chen, X.~He, H.~Li, X.~Jin, J.~Feng, and J.~Liu, ``Cosa: Concatenated sample pretrained vision-language foundation model,'' \emph{arXiv preprint arXiv:2306.09085}, 2023.

\bibitem{wang2022git}
J.~Wang, Z.~Yang, X.~Hu, L.~Li, K.~Lin, Z.~Gan, Z.~Liu, C.~Liu, and L.~Wang, ``Git: A generative image-to-text transformer for vision and language,'' \emph{arXiv preprint arXiv:2205.14100}, 2022.

\bibitem{wang2022image}
W.~Wang, H.~Bao, L.~Dong, J.~Bjorck, Z.~Peng, Q.~Liu, K.~Aggarwal, O.~K. Mohammed, S.~Singhal, S.~Som \emph{et~al.}, ``Image as a foreign language: Beit pretraining for all vision and vision-language tasks,'' \emph{arXiv preprint arXiv:2208.10442}, 2022.

\bibitem{bao2022vlmo}
H.~Bao, W.~Wang, L.~Dong, Q.~Liu, O.~K. Mohammed, K.~Aggarwal, S.~Som, S.~Piao, and F.~Wei, ``Vlmo: Unified vision-language pre-training with mixture-of-modality-experts,'' \emph{Advances in Neural Information Processing Systems}, vol.~35, pp. 32\,897--32\,912, 2022.

\bibitem{hao2022language}
Y.~Hao, H.~Song, L.~Dong, S.~Huang, Z.~Chi, W.~Wang, S.~Ma, and F.~Wei, ``Language models are general-purpose interfaces,'' \emph{arXiv preprint arXiv:2206.06336}, 2022.

\bibitem{huang2023language}
S.~Huang, L.~Dong, W.~Wang, Y.~Hao, S.~Singhal, S.~Ma, T.~Lv, L.~Cui, O.~K. Mohammed, Q.~Liu \emph{et~al.}, ``Language is not all you need: Aligning perception with language models,'' \emph{arXiv preprint arXiv:2302.14045}, 2023.

\bibitem{peng2023kosmos}
Z.~Peng, W.~Wang, L.~Dong, Y.~Hao, S.~Huang, S.~Ma, and F.~Wei, ``Kosmos-2: Grounding multimodal large language models to the world,'' \emph{arXiv preprint arXiv:2306.14824}, 2023.

\bibitem{lu2022unified}
J.~Lu, C.~Clark, R.~Zellers, R.~Mottaghi, and A.~Kembhavi, ``Unified-io: A unified model for vision, language, and multi-modal tasks,'' \emph{arXiv preprint arXiv:2206.08916}, 2022.

\bibitem{xu2023mplug}
H.~Xu, Q.~Ye, M.~Yan, Y.~Shi, J.~Ye, Y.~Xu, C.~Li, B.~Bi, Q.~Qian, W.~Wang \emph{et~al.}, ``mplug-2: A modularized multi-modal foundation model across text, image and video,'' \emph{arXiv preprint arXiv:2302.00402}, 2023.

\bibitem{kuo2023mammut}
W.~Kuo, A.~Piergiovanni, D.~Kim, X.~Luo, B.~Caine, W.~Li, A.~Ogale, L.~Zhou, A.~Dai, Z.~Chen \emph{et~al.}, ``Mammut: A simple architecture for joint learning for multimodal tasks,'' \emph{arXiv preprint arXiv:2303.16839}, 2023.

\bibitem{hu2022promptcap}
Y.~Hu, H.~Hua, Z.~Yang, W.~Shi, N.~A. Smith, and J.~Luo, ``Promptcap: Prompt-guided task-aware image captioning,'' \emph{arXiv preprint arXiv:2211.09699}, 2022.

\bibitem{guo2022images}
J.~Guo, J.~Li, D.~Li, A.~M.~H. Tiong, B.~Li, D.~Tao, and S.~C. Hoi, ``From images to textual prompts: Zero-shot vqa with frozen large language models,'' \emph{arXiv preprint arXiv:2212.10846}, 2022.

\bibitem{zhang2023moqagpt}
L.~Zhang, Y.~Wu, F.~Mo, J.-Y. Nie, and A.~Agrawal, ``Moqagpt: Zero-shot multi-modal open-domain question answering with large language model,'' \emph{arXiv preprint arXiv:2310.13265}, 2023.

\bibitem{chen2023vast}
S.~Chen, H.~Li, Q.~Wang, Z.~Zhao, M.~Sun, X.~Zhu, and J.~Liu, ``Vast: A vision-audio-subtitle-text omni-modality foundation model and dataset,'' \emph{Advances in Neural Information Processing Systems}, vol.~36, pp. 72\,842--72\,866, 2023.

\bibitem{Radford2021LearningTV}
A.~Radford, J.~W. Kim, C.~Hallacy, A.~Ramesh, G.~Goh, S.~Agarwal, G.~Sastry, A.~Askell, P.~Mishkin, J.~Clark, G.~Krueger, and I.~Sutskever, ``Learning transferable visual models from natural language supervision,'' in \emph{International Conference on Machine Learning}, 2021.

\bibitem{lu2022learn}
P.~Lu, S.~Mishra, T.~Xia, L.~Qiu, K.-W. Chang, S.-C. Zhu, O.~Tafjord, P.~Clark, and A.~Kalyan, ``Learn to explain: Multimodal reasoning via thought chains for science question answering,'' \emph{Advances in Neural Information Processing Systems}, vol.~35, pp. 2507--2521, 2022.

\bibitem{Khashabi2020UnifiedQACF}
D.~Khashabi, S.~Min, T.~Khot, A.~Sabharwal, O.~Tafjord, P.~Clark, and H.~Hajishirzi, ``Unifiedqa: Crossing format boundaries with a single qa system,'' in \emph{Findings}, 2020.

\bibitem{gao2023assistgpt}
D.~Gao, L.~Ji, L.~Zhou, K.~Q. Lin, J.~Chen, Z.~Fan, and M.~Z. Shou, ``Assistgpt: A general multi-modal assistant that can plan, execute, inspect, and learn,'' \emph{arXiv preprint arXiv:2306.08640}, 2023.

\bibitem{khot2022decomposed}
T.~Khot, H.~Trivedi, M.~Finlayson, Y.~Fu, K.~Richardson, P.~Clark, and A.~Sabharwal, ``Decomposed prompting: A modular approach for solving complex tasks,'' \emph{arXiv preprint arXiv:2210.02406}, 2022.

\bibitem{huang2023instruct2act}
S.~Huang, Z.~Jiang, H.~Dong, Y.~Qiao, P.~Gao, and H.~Li, ``Instruct2act: Mapping multi-modality instructions to robotic actions with large language model,'' \emph{arXiv preprint arXiv:2305.11176}, 2023.

\bibitem{suris2023vipergpt}
D.~Sur{\'\i}s, S.~Menon, and C.~Vondrick, ``Vipergpt: Visual inference via python execution for reasoning,'' \emph{arXiv preprint arXiv:2303.08128}, 2023.

\bibitem{choudhury2023zero}
R.~Choudhury, K.~Niinuma, K.~M. Kitani, and L.~A. Jeni, ``Zero-shot video question answering with procedural programs,'' \emph{arXiv preprint arXiv:2312.00937}, 2023.

\bibitem{gao2024cantor}
T.~Gao, P.~Chen, M.~Zhang, C.~Fu, Y.~Shen, Y.~Zhang, S.~Zhang, X.~Zheng, X.~Sun, L.~Cao \emph{et~al.}, ``Cantor: Inspiring multimodal chain-of-thought of mllm,'' \emph{arXiv preprint arXiv:2404.16033}, 2024.

\bibitem{chen2023video}
J.~Chen, D.~Zhu, K.~Haydarov, X.~Li, and M.~Elhoseiny, ``Video chatcaptioner: Towards the enriched spatiotemporal descriptions,'' \emph{arXiv preprint arXiv:2304.04227}, 2023.

\bibitem{ahn2022can}
M.~Ahn, A.~Brohan, N.~Brown, Y.~Chebotar, O.~Cortes, B.~David, C.~Finn, C.~Fu, K.~Gopalakrishnan, K.~Hausman \emph{et~al.}, ``Do as i can, not as i say: Grounding language in robotic affordances,'' \emph{arXiv preprint arXiv:2204.01691}, 2022.

\bibitem{yang2022empirical}
Z.~Yang, Z.~Gan, J.~Wang, X.~Hu, Y.~Lu, Z.~Liu, and L.~Wang, ``An empirical study of gpt-3 for few-shot knowledge-based vqa,'' in \emph{Proceedings of the AAAI Conference on Artificial Intelligence}, vol.~36, no.~3, 2022, pp. 3081--3089.

\bibitem{caron2021emerging}
M.~Caron, H.~Touvron, I.~Misra, H.~J{\'e}gou, J.~Mairal, P.~Bojanowski, and A.~Joulin, ``Emerging properties in self-supervised vision transformers,'' in \emph{Proceedings of the IEEE/CVF international conference on computer vision}, 2021, pp. 9650--9660.

\bibitem{ramesh2021zero}
A.~Ramesh, M.~Pavlov, G.~Goh, S.~Gray, C.~Voss, A.~Radford, M.~Chen, and I.~Sutskever, ``Zero-shot text-to-image generation,'' in \emph{International Conference on Machine Learning}.\hskip 1em plus 0.5em minus 0.4em\relax PMLR, 2021, pp. 8821--8831.

\bibitem{shao2023prompting}
Z.~Shao, Z.~Yu, M.~Wang, and J.~Yu, ``Prompting large language models with answer heuristics for knowledge-based visual question answering,'' in \emph{Proceedings of the IEEE/CVF Conference on Computer Vision and Pattern Recognition}, 2023, pp. 14\,974--14\,983.

\bibitem{zhou2023prompting}
Y.~Zhou, P.~Cao, Y.~Chen, K.~Liu, and J.~Zhao, ``Prompting vision language model with knowledge from large language model for knowledge-based vqa,'' \emph{arXiv preprint arXiv:2308.15851}, 2023.

\bibitem{nakano2021webgpt}
R.~Nakano, J.~Hilton, S.~Balaji, J.~Wu, L.~Ouyang, C.~Kim, C.~Hesse, S.~Jain, V.~Kosaraju, W.~Saunders \emph{et~al.}, ``Webgpt: Browser-assisted question-answering with human feedback,'' \emph{arXiv preprint arXiv:2112.09332}, 2021.

\bibitem{yu2022generate}
W.~Yu, D.~Iter, S.~Wang, Y.~Xu, M.~Ju, S.~Sanyal, C.~Zhu, M.~Zeng, and M.~Jiang, ``Generate rather than retrieve: Large language models are strong context generators,'' \emph{arXiv preprint arXiv:2209.10063}, 2022.

\bibitem{liu2023ml}
Y.~Liu, X.~Tang, Z.~Cai, J.~Lu, Y.~Zhang, Y.~Shao, Z.~Deng, H.~Hu, Z.~Yang, K.~An \emph{et~al.}, ``Ml-bench: Large language models leverage open-source libraries for machine learning tasks,'' \emph{arXiv preprint arXiv:2311.09835}, 2023.

\bibitem{xiang2023multi}
C.~Xiang, C.~Feng, X.~Xie, B.~Shi, H.~Lu, Y.~Lv, M.~Yang, and Z.~Niu, ``Multi-sensor fusion and cooperative perception for autonomous driving: A review,'' \emph{IEEE Intelligent Transportation Systems Magazine}, 2023.

\bibitem{huang2022multi}
K.~Huang, B.~Shi, X.~Li, X.~Li, S.~Huang, and Y.~Li, ``Multi-modal sensor fusion for auto driving perception: A survey,'' \emph{arXiv preprint arXiv:2202.02703}, 2022.

\bibitem{jiao2023msmdfusion}
Y.~Jiao, Z.~Jie, S.~Chen, J.~Chen, L.~Ma, and Y.-G. Jiang, ``Msmdfusion: Fusing lidar and camera at multiple scales with multi-depth seeds for 3d object detection,'' in \emph{Proceedings of the IEEE/CVF conference on computer vision and pattern recognition}, 2023, pp. 21\,643--21\,652.

\bibitem{li2022voxel}
Y.~Li, X.~Qi, Y.~Chen, L.~Wang, Z.~Li, J.~Sun, and J.~Jia, ``Voxel field fusion for 3d object detection,'' in \emph{Proceedings of the IEEE/CVF Conference on Computer Vision and Pattern Recognition}, 2022, pp. 1120--1129.

\bibitem{bang2023multitask}
Y.~Bang, S.~Cahyawijaya, N.~Lee, W.~Dai, D.~Su, B.~Wilie, H.~Lovenia, Z.~Ji, T.~Yu, W.~Chung \emph{et~al.}, ``A multitask, multilingual, multimodal evaluation of chatgpt on reasoning, hallucination, and interactivity,'' \emph{arXiv preprint arXiv:2302.04023}, 2023.

\bibitem{li2023evaluating}
Y.~Li, Y.~Du, K.~Zhou, J.~Wang, W.~X. Zhao, and J.-R. Wen, ``Evaluating object hallucination in large vision-language models,'' \emph{arXiv preprint arXiv:2305.10355}, 2023.

\bibitem{glaese2022improving}
A.~Glaese, N.~McAleese, M.~Trebacz, J.~Aslanides, V.~Firoiu, T.~Ewalds, M.~Rauh, L.~Weidinger, M.~Chadwick, P.~Thacker \emph{et~al.}, ``Improving alignment of dialogue agents via targeted human judgements,'' \emph{arXiv preprint arXiv:2209.14375}, 2022.

\bibitem{sun2024trustllm}
L.~Sun, Y.~Huang, H.~Wang, S.~Wu, Q.~Zhang, C.~Gao, Y.~Huang, W.~Lyu, Y.~Zhang, X.~Li \emph{et~al.}, ``Trustllm: Trustworthiness in large language models,'' \emph{arXiv preprint arXiv:2401.05561}, 2024.

\bibitem{singh2024rethinking}
C.~Singh, J.~P. Inala, M.~Galley, R.~Caruana, and J.~Gao, ``Rethinking interpretability in the era of large language models,'' \emph{arXiv preprint arXiv:2402.01761}, 2024.

\bibitem{creswell2022selection}
A.~Creswell, M.~Shanahan, and I.~Higgins, ``Selection-inference: Exploiting large language models for interpretable logical reasoning,'' \emph{arXiv preprint arXiv:2205.09712}, 2022.

\bibitem{xue2024few}
D.~Xue, S.~Qian, and C.~Xu, ``Few-shot multimodal explanation for visual question answering,'' in \emph{ACM Multimedia 2024}, 2024.

\bibitem{bai2024beyond}
G.~Bai, Z.~Chai, C.~Ling, S.~Wang, J.~Lu, N.~Zhang, T.~Shi, Z.~Yu, M.~Zhu, Y.~Zhang \emph{et~al.}, ``Beyond efficiency: A systematic survey of resource-efficient large language models,'' \emph{arXiv preprint arXiv:2401.00625}, 2024.

\bibitem{ford2021portauthority}
B.~W. Ford and Z.~Zong, ``Portauthority: Integrating energy efficiency analysis into cross-platform development cycles via dynamic program analysis,'' \emph{Sustainable Computing: Informatics and Systems}, vol.~30, p. 100530, 2021.

\bibitem{xie2019integration}
Y.~Xie, S.~Chen, Q.~Ni, and H.~Wu, ``Integration of resource allocation and task assignment for optimizing the cost and maximum throughput of business processes,'' \emph{Journal of Intelligent Manufacturing}, vol.~30, pp. 1351--1369, 2019.

\bibitem{wang2024model}
W.~Wang, W.~Chen, Y.~Luo, Y.~Long, Z.~Lin, L.~Zhang, B.~Lin, D.~Cai, and X.~He, ``Model compression and efficient inference for large language models: A survey,'' \emph{arXiv preprint arXiv:2402.09748}, 2024.

\end{thebibliography}

\newpage

\vfill

\end{document}